\documentclass{article} 
\usepackage{iclr2027_conference,times}

\usepackage{amsmath,amsfonts,bm}

\def\eqref#1{equation~\ref{#1}}

\def\1{\bm{1}}

\DeclareMathAlphabet{\mathsfit}{\encodingdefault}{\sfdefault}{m}{sl}
\SetMathAlphabet{\mathsfit}{bold}{\encodingdefault}{\sfdefault}{bx}{n}

\usepackage{hyperref}
\usepackage{url}
\usepackage[utf8]{inputenc} 
\usepackage[T1]{fontenc}    
\usepackage{booktabs}       
\usepackage{amsfonts}       
\usepackage{nicefrac}       
\usepackage{microtype}      
\usepackage{xcolor}         
\usepackage{graphicx}
\usepackage{enumitem}
\usepackage{multirow}
\usepackage{amsmath}
\usepackage{makecell}
\usepackage{wrapfig}
\usepackage[title]{appendix}
\definecolor{BurntOrange}{cmyk}{0, 0.51, 1, 0}
\usepackage{tabularx}
\usepackage{array}
\usepackage{ragged2e}
\usepackage{graphicx}
\usepackage{booktabs}
\usepackage{capt-of}

\title{UHR-Micro: Diagnosing and Mitigating the Resolution Illusion in Earth Observation VLMs}

\author{
\normalfont
Shuo Ni$^{1,3\star}$,
Tong Wang$^{1\star}$,
Jing Zhang$^{2,3\dagger}$,
Di Wang$^{2,3}$,
He Chen$^{1}$,
Haonan Guo$^{3,4}$,\\
\normalfont
Ning Zhang$^{1,5\dagger}$,
Bo Du$^{2,3\dagger}$\\[2pt]
\normalfont
$^{1}$National Key Laboratory of Science and Technology on Space-Born Intelligent Information \\Processing, Beijing Institute of Technology, Beijing, China \\
\normalfont
$^{2}$School of Computer Science, Wuhan University, Wuhan, China \\
\normalfont
$^{3}$Zhongguancun Academy, Beijing, China\\
\normalfont
$^4$ State Key Laboratory of Information Engineering in Surveying, Mapping and Remote Sensing,\\ Wuhan University, Wuhan, China\\
\normalfont
$^{5}$The Department of Computing, Hong Kong Polytechnic University, Hong Kong\\[2pt]
\small
\texttt{shuoni@bit.edu.cn};
\texttt{jingzhang.cv@gmail.com}\\
\small
\texttt{nzhang.rs@bit.edu.cn};
\texttt{dubo@whu.edu.cn}\\[2pt]
\\
$^\star$ Equal contribution. $^\dagger$ Corresponding authors.
}

\iclrfinalcopy 
\begin{document}

\maketitle

\lhead{Preprint. Under review.}

\begin{abstract}
  Vision-Language Models (VLMs) are increasingly used to analyze ultra-high-resolution (UHR) Earth observation imagery, yet they face a severe scale mismatch between broad scene context and micro-scale targets. We refer to this phenomenon as a “resolution illusion”: higher input resolution provides access to more visual detail, but does not necessarily translate into reliable perception of task-relevant micro-evidence. To benchmark this challenge, we introduce UHR-Micro, a benchmark comprising 11,072 instructions grounded in 1,212 UHR images, designed to evaluate VLMs on micro-scale evidence in native-resolution Earth observation imagery. UHR-Micro spans diverse target scales, task families, and visual conditions, with each sample having an objectively verifiable target. Experiments with representative high-resolution VLMs show substantial failures in localizing and interpreting task-relevant evidence, despite access to high-resolution inputs. Further analysis shows that model scaling alone is insufficient, while localized evidence substantially improves performance, pointing to evidence access as a major bottleneck. Motivated by this finding, we propose Micro-evidence Active Perception (MAP), which constructs a spatially grounded evidence state from localized observations and supplements it when the current evidence is insufficient. Across two backbone VLMs, MAP improves UHR-Micro performance by 8.65 percentage points on average. UHR-Micro and MAP provide a framework for diagnosing and improving high-resolution reasoning in Earth observation VLMs. Datasets and source code were released at \href{https://github.com/MiliLab/UHR-Micro}{\color{magenta}UHR-Micro}.
\end{abstract}

\section{Introduction}
Recent advances in Vision-Language Models (VLMs) have substantially improved general visual understanding~\citep{zhang2024vision,zhou2022learning}. Yet applying these models to Earth observation exposes a fundamental scale mismatch between large-scale geographic contexts and micro-targets~\citep{danish2025geobench,ni2025unigeoseg}. Ultra-high-resolution (UHR) remote sensing images often span broad scenes, while task-relevant evidence may occupy only a minuscule fraction of the pixels~\citep{wang2025xlrs, yang2025advances}. Such evidence can include vehicles, vessels, small facilities, or narrow infrastructure~\citep{wang2021aitod}. This extreme scale disparity requires VLMs to preserve global scene context while performing fine-grained perception~\citep{miri2025soda}.

\begin{figure}[t]
  \centering
  \includegraphics[width=0.9\linewidth]{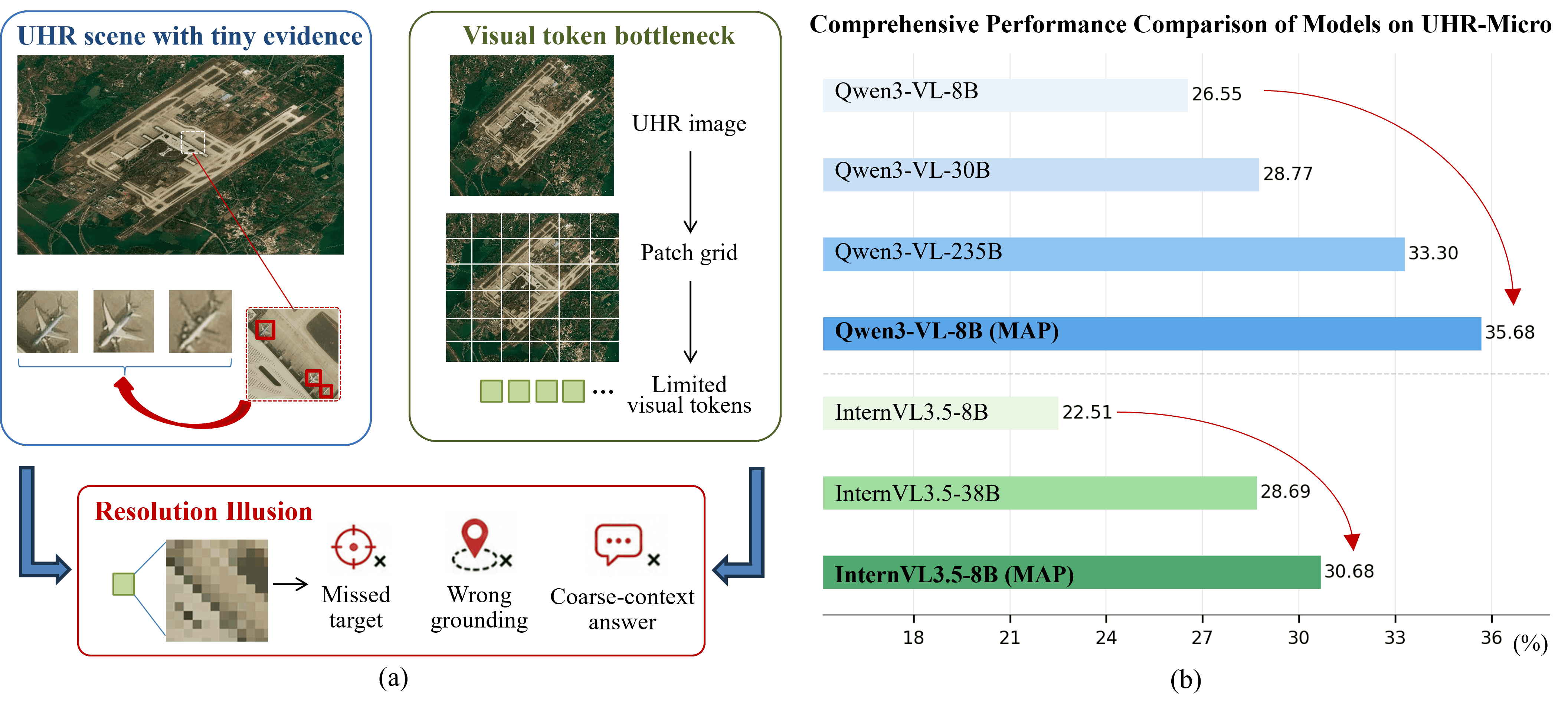}
\caption{
Resolution illusion and MAP.
\textbf{(a)} UHR images contain fine-grained evidence that can be difficult for VLMs to access under limited visual representations, encouraging reliance on coarse context.
\textbf{(b)} MAP actively acquires and supplements localized evidence while preserving its spatial grounding, improving UHR-Micro performance over backbone models.
}
  \label{fig1}
\end{figure}

Standard VLM pipelines are usually constrained by bounded visual-token budgets. When UHR images are resized~\citep{liu2024improvedllave}, patched~\citep{li2024llavanext}, or compressed into a limited set of visual tokens~\citep{chen2024internvl}, small visual cues can become difficult to preserve or access during visual processing~\citep{zhang2024mme,shabbir2025geopixel}. We refer to this empirical gap as a \emph{resolution illusion}: higher input resolution provides access to more visual detail, but does not necessarily translate into reliable perception of task-relevant micro-evidence. In Earth observation, such failures can lead to missed targets, incorrect spatial grounding, or answers based on coarse scene context rather than localized observations, as shown in Fig.~\ref{fig1}.

Existing remote sensing VLM benchmarks do not systematically isolate this failure mode. Most focus on image-level semantics~\citep{lu2017ucmcaption,kuckreja2024geochat}, salient objects~\citep{sun2022rsvg,zhan2023rsvgdior}, or coarse scene understanding~\citep{muhtar2024lhrs,an2024choice,li2024vrsbench}. These tasks are valuable, but they do not test whether a model can locate and use micro-evidence in large UHR scenes. They also offer limited diagnostic control over target scale, spatial distribution, context range, task family, and visual conditions~\citep{liu2026seeing}. Without such diagnostic control, it is difficult to distinguish genuine micro-evidence grounding from answers driven by coarse visual context, language priors, or dataset shortcuts~\citep{dang2025benchmarkUHR}. A dedicated benchmark is therefore needed to evaluate UHR reasoning at the level of task-relevant micro-evidence.

To address this gap, we introduce \textbf{UHR-Micro}, a benchmark for evaluating whether VLMs can locate, interpret, and reason over task-relevant micro-evidence in UHR Earth observation imagery. UHR-Micro contains 11,072 instructions paired with 1,212 curated UHR images, with task-relevant targets occupying less than $0.01\%$ of the image area on average. These instructions span 16 tasks across diverse target sizes, context requirements, visual conditions, and four task families: grounding, counting, spatial reasoning, and fine-grained understanding. Each instruction is centered on micro-evidence that must be identified and used to answer the query, enabling controlled evaluation of the gap between nominal input resolution and effective evidence grounding.

UHR-Micro is constructed through a hybrid human-in-the-loop annotation process designed to ensure that each instruction depends on verifiable visual evidence rather than language-only priors. Experiments on representative high-resolution VLMs reveal substantial failures in locating and interpreting relevant visual evidence, even when models have access to high-resolution inputs. Further analysis of these experiments suggests that increasing model capacity provides only partial gains, whereas supplying localized native-resolution evidence substantially improves performance, pointing to evidence access as a major bottleneck.

Motivated by this finding, we propose \textbf{Micro-evidence Active Perception (MAP)}, a training-free method that constructs and refines a spatially grounded evidence state for high-resolution visual reasoning. MAP selectively acquires task-relevant observations, preserves their spatial correspondence, and completes the evidence state when observations are insufficient. Across two backbones, MAP improves UHR-Micro performance by 8.65 percentage points on average over direct inference baselines, supporting evidence-centered perception as an effective strategy for UHR reasoning.

Our main contributions are summarized as follows:
\begin{itemize}[leftmargin=*, topsep=0pt]
    \item \textbf{UHR-Micro Benchmark.} We introduce UHR-Micro, a diagnostic benchmark comprising 11,072 instructions grounded in 1,212 UHR Earth observation images, spanning 16 tasks centered on task-relevant micro-evidence, with targets occupying less than $0.01\%$ of the image area on average.

    \item \textbf{Resolution Illusion Analysis.} Experimental analysis on UHR-Micro characterizes the \emph{resolution illusion} as the gap between nominal input resolution and effective micro-evidence grounding, revealing failures in spatial grounding and evidence interpretation across representative VLMs.

    \item \textbf{Micro-evidence Active Perception.} We propose MAP, a training-free method that constructs a spatially grounded evidence state from localized observations and refines it when additional evidence is required, improving UHR-Micro performance by 8.65 percentage points on average across two backbones.

\end{itemize}

\begin{figure}[t]
  \centering
  \includegraphics[width=\linewidth]{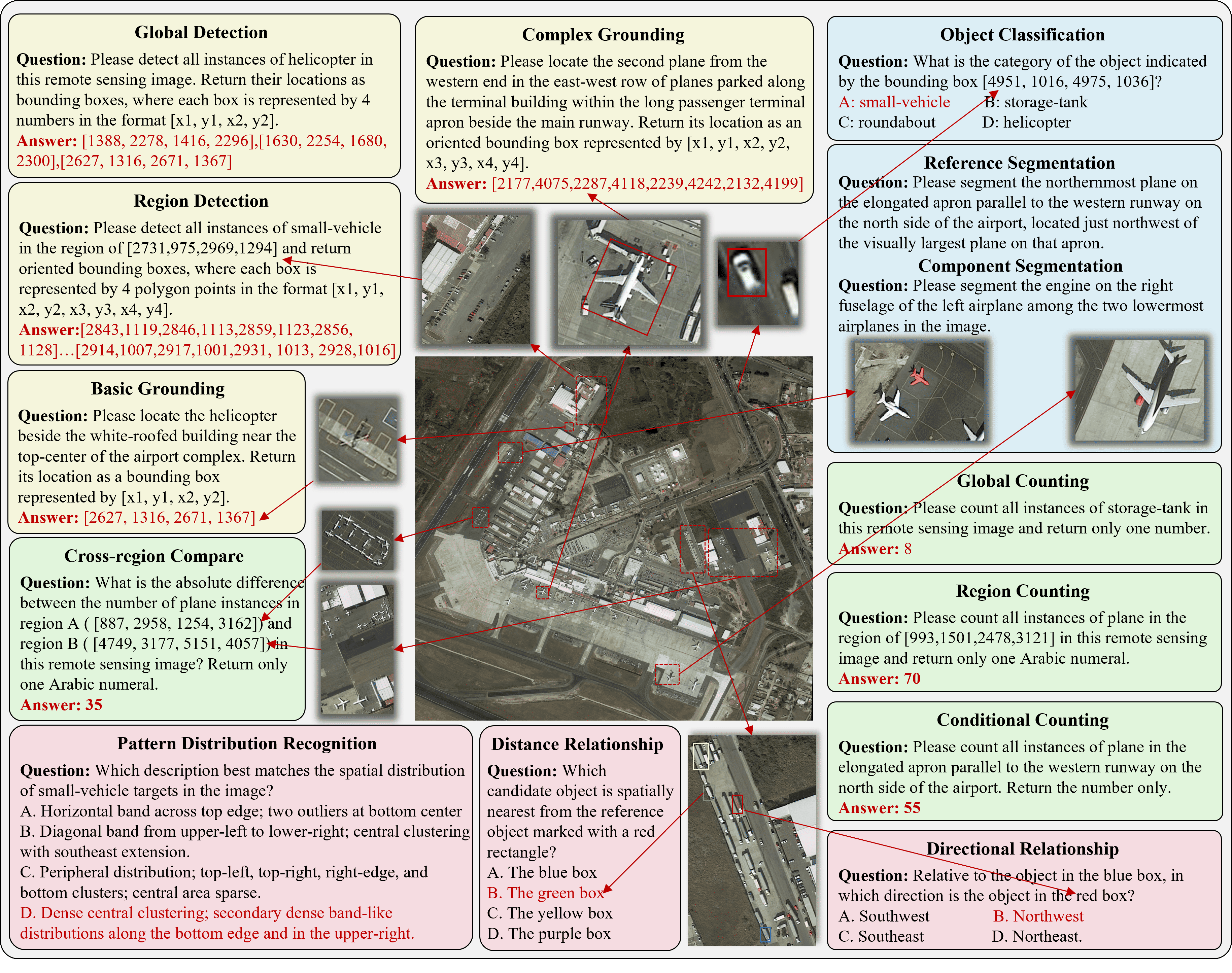}
\caption{
Overview of UHR-Micro.
The benchmark evaluates micro-level perception in UHR Earth observation across grounding, fine-grained understanding, counting, and spatial reasoning.
Each task is centered on spatially tiny evidence, forcing models to locate and use local observations rather than relying on coarse scene context.
}
  \label{fig2}
\end{figure}

\section{Related Work}
\textbf{Multimodal Benchmarks in Earth Observation.} Multimodal remote sensing evaluation initially centered on task-specific datasets, with RSVQA-HR \citep{lobry2020rsvqa} and RSVG \citep{sun2022rsvg} establishing paradigms for visual question answering and region grounding. The field has since moved toward broader instruction-following benchmarks: LHRS-Bench \citep{muhtar2024lhrs} introduced multi-granularity tasks, while VRSBench \citep{li2024vrsbench} targeted region-level visual comprehension at scale. Recent efforts further scale evaluation through zero-shot diagnostics in GEOBench-VLM \citep{danish2025geobench} and globally distributed data in OmniEarth-Bench \citep{wang2026omniearthbenchholisticevaluationearths}. However, extremely small objects remain systematically underexplored. Existing benchmarks often rely on downsampling, cropped patches, or macro-level semantics even for large scenes \citep{wang2025xlrs}, bypassing the severe scale disparity of native UHR imagery. UHR-Micro fills this gap by evaluating micro-level perception directly in UHR scenes.

\textbf{Resolution Constraints and Perceptual Limitations.} Fixed-resolution encoders such as CLIP \citep{radford2021CLIP} inevitably erase micro-features when resizing ultra-high-resolution imagery. Although recent models adopt dynamic strategies, including adaptive tiling in LLaVA-NeXT \citep{li2024llavanext} and InternVL3.5 \citep{wang2025internvl3}, or dynamic resolution encoding in Qwen3-VL \citep{bai2025qwen3}, they still face fundamental perceptual bottlenecks. Minute object signals may be lost during token compression or overwhelmed by background noise, causing models to rely on linguistic priors and hallucinate \citep{li2023evaluating}. UHR-Micro operationalizes this ``resolution illusion'' by measuring whether nominal high-resolution input access translates into effective micro-level perception.

\textbf{Strategic Inference and Active Perception.}
Active perception has been explored to improve visual reasoning under limited visual budgets. DeepEyes~\citep{zheng2026deepeyes} adaptively crops informative regions, enabling VLMs to inspect fine-grained details beyond a global view. In Earth observation, GeoEyes~\citep{wang2026geoeyes} and ZoomEarth~\citep{liu2025zoomearth} incorporate VLM agents, region selection, or iterative zooming for geospatial reasoning. These works demonstrate the value of localized inspection for high-resolution understanding. MAP follows this line of work, but organizes localized observations into a spatially grounded evidence state and selectively supplements missing evidence when needed.

\section{UHR-Micro}
\subsection{Task Taxonomy for Micro-Level Perception}
To systematically probe the perceptual limits of VLMs under extreme scale disparities, we construct a evaluation taxonomy organized around four dimensions. Each dimension isolates a fundamental facet of visual perception. Together, they systematically evaluate not only where and what the micro-targets are, but also how many exist and how they spatially relate across the full image, providing a holistic assessment of micro-level perception.

\begin{table}
\caption{
Dataset statistics and comparison with representative vision-language benchmarks. RS denotes remote sensing, FF denotes free-form answers.
}
\label{tab_datasets}
\centering
\resizebox{\textwidth}{!}{%
\begin{tabular}{lcccccc}
\toprule
Dataset & Domain & Average Resolution & Average Target Size & Target Area Proportion & Volume & Output Type \\
\midrule
MMBench~\citep{liu2024mmbench}        & General & $512\times270$     & -               & -        & 3,217     & Option \\
MMstar~\citep{chen2024mmstar}         & General & $512\times375$     & -               & -        & 1,500     & Option \\
RefCOCO~\citep{yu2016refcoco}        & General & $640\times480$     & $70\times70$    & 1.59\%   & 17,596    & BBox \\
MME-Realworld~\citep{zhang2024mme}  & General & $2,000\times1,500$ & -               & -        & 29,429    & Option \\
UCM-Caption~\citep{lu2017ucmcaption}    & RS      & $256\times256$     & -               & -        & 10,500    & FF \\
RSVQA-HR~\citep{lobry2020rsvqa}       & RS      & $1,024\times1,024$ & -               & -        & 1,066,316 & Option \\
RSVG~\citep{sun2022rsvg}           & RS      & $800\times800$     & $60\times60$    & 0.56\%   & 7,933     & BBox \\
DIOR-RSVG~\citep{zhan2023rsvgdior}      & RS      & $800\times800$     & $170\times170$  & 4.52\%   & 38,320    & BBox \\
RRSIS-D~\citep{liu2024rrsisd}        & RS      & $800\times800$     & $200\times200$  & 6.25\%   & 17,402    & Mask \\
LHRS-Bench~\citep{muhtar2024lhrs}     & RS      & $460\times420$     & -               & -        & 140,000   & Option \\
Geochat-bench~\citep{kuckreja2024geochat}  & RS      & $630\times620$     & $120\times120$  & 3.69\%   & 318,000   & FF, BBox \\
CHOICE~\citep{an2024choice}         & RS      & $650\times650$     & $160\times130$  & 4.92\%   & 9,550     & Option, BBox, Mask \\
VRSBench~\citep{li2024vrsbench}       & RS      & $512\times512$     & $120\times120$  & 5.49\%   & 205,317   & FF, BBox \\
XLRS-Bench~\citep{wang2025xlrs}     & RS      & $8,500\times8,000$ & $200\times200$  & 0.06\%    & 45,942    & FF, Option, BBox \\
\midrule
UHR-Micro      & RS      & $4,800\times3,200$ & $50\times30$  & \textbf{0.01\%}  & 11,072    & Option, Number, BBox, Mask \\
\bottomrule
\end{tabular}}
\end{table}

\textbf{Grounding.} Focused on spatial anchoring, this dimension evaluates the model's ability to locate micro-targets amid vast backgrounds through five tasks. \textit{Global Detection (GD)} requires search over the entire image, while \textit{Regional Detection (RD)} restricts detection to bounded sub-areas. \textit{Basic Grounding (BG)} identifies unique targets from simple spatial cues without similar distractors, whereas \textit{Complex Grounding (CG)} requires multi-step reasoning over spatial and relational cues, such as exact ordinal positioning within dense clusters. \textit{Multi-Condition Retrieval (MCR)} further asks models to retrieve all targets satisfying combined textual attributes and spatial relations.

\textbf{Fine-grained Understanding.} Beyond localization, fine-grained understanding examines whether models can capture semantic and morphological details of micro-targets. \textit{Object Classification (OC)} tests category assignment for extremely small instances. \textit{Fine-Grained Recognition (FGR)} requires identifying specific subtypes, such as aircraft or ship models. At the pixel level, \textit{Referring Segmentation (RS)} is evaluated against object masks specified by textual descriptions, while \textit{Component Segmentation (CS)} is evaluated against masks of tiny components such as aircraft empennages.

\textbf{Counting.} For numerical perception, UHR-Micro evaluates models on dense micro-target enumeration through four tasks. \textit{Global Counting (GC)} and \textit{Regional Counting (RC)} require exhaustive counting over the full image and bounded regions, respectively. \textit{Conditional Counting (CC)} introduces semantic constraints by counting targets with specified visual attributes. \textit{Cross-Region Compare (CRC)} further requires tracking disjoint spatial regions and computing numerical differences between their target populations.

\textbf{Spatial Reasoning.} To assess geometric and topological awareness, we include three spatial reasoning tasks over micro-targets. \textit{Directional Relationship (DrR)} and \textit{Distance Relationship (DsR)} test local pairwise spatial relations, while \textit{Pattern Distribution Recognition (PDR)} examines global distribution patterns and object-swarm topology.

To avoid subjective bias, all tasks are unified into four deterministic output formats: bounding boxes, segmentation masks, exact numbers, and discrete options. This design ensures that the benchmark remains unambiguous, quantitative, and fully reproducible.

\begin{table}[]
\caption{Performance on the UHR-Micro validation split for Qwen3-VL-8B and InternVL3.5-8B under different UHR perception strategies. Scores are percentages. Dashes denote strategy--task pairs where no separate score is reported; when computing Avg.$^\dagger$, these entries are filled with the corresponding Native score of the same model and task to preserve a common task distribution.}
\label{tab_val}
\centering
\resizebox{\textwidth}{!}{%
\begin{tabular}{lc|ccccc|cccc|cccc|ccc|c}
\toprule
& & \multicolumn{5}{c|}{Grounding} & \multicolumn{4}{c|}{Fine-grained Understanding} & \multicolumn{4}{c|}{Counting} & \multicolumn{3}{c|}{Spatial Reasoning} & \\
\multirow{-2}{*}{Model} & \multicolumn{1}{c|}{\multirow{-2}{*}{Strategy}} & GD & RD & BG & CG & MCR & OC & FGR & RS & CS & GC & RC & CC & CRC & DrR & DsR & PDR & \multicolumn{1}{c}{\multirow{-2}{*}{Avg.$^\dagger$}} \\
\midrule
 & Native & 2.10 & 0.47 & 16.33 & 5.55 & 4.22 & 76.53 & 29 & 13.77 & 6.29 & 23.40 & 12.87 & 12.80 & 14.56 & 75 & 76 & 74 & 27.68 \\
 & Resize-1024 & 0.25 & 0.79 & 9.11 & 4.20 & 2.58 & 59.18 & 28 & 7.58 & 4.08 & 12.80 & 4.40 & 11.70 & 16.49 & 60 & 57 & 65 & 21.45 \\
 & Query Crop & - & 4.49 & - & - & - & 58.16 & 37 & - & 6.58 & - & 36.68 & - & - & - & - & - & 28.79 \\
 & Oracle GT-Crop-640 & - & 1.27 & 48.24 & 41.15 & - & 87.76 & 35 & - & - & - & 32.52 & - & - & - & - & - & 34.26 \\
 & Oracle GT-Crop-1024 & - & 1.76 & 42.68 & 29.22 & - & 86.73 & 31 & - & - & - & 31.02 & - & - & - & - & - & 32.78 \\
 & Oracle GT-Crop-2048 & - & 1.38 & 28.00 & 17.42 & - & 78.57 & 28 & - & - & - & 24.00 & - & - & - & - & - & 29.97 \\
\multirow{-7}{*}{Qwen3-VL-8B} & Sliding Window & 7.87 & 0.07 & 15.55 & 6.95 & 4.05 & 67.35 & 27 & 13.71 & 4.34 & 33.31 & 9.70 & 13.08 & - & - & - & - & 27.66 \\
\midrule
 & Native & 2.29 & 0.90 & 6.65 & 2.50 & 2.76 & 59.18 & 26 & 9.46 & 2.32 & 43.30 & 30.37 & 28.52 & 31.54 & 39 & 39 & 45 & 23.05 \\
 & Resize-1024 & 3.38 & 0.66 & 7.48 & 2.50 & 3.25 & 62.24 & 26 & 8.70 & 3.27 & 38.30 & 23.12 & 26.62 & 28.59 & 39 & 38 & 54 & 22.82 \\
 & Query Crop & - & 20.19 & - & - & - & 61.22 & 44 & - & 1.83 & - & 52.29 & - & - & - & - & - & 26.85 \\
 & Oracle GT-Crop-640 & - & 17.88 & 38.37 & 34.90 & - & 85.71 & 46 & - & - & - & 52.66 & - & - & - & - & - & 32.42 \\
 & Oracle GT-Crop-1024 & - & 12.86 & 33.86 & 30.63 & - & 81.63 & 45 & - & - & - & 48.66 & - & - & - & - & - & 30.99 \\
 & Oracle GT-Crop-2048 & - & 4.34 & 18.12 & 16.46 & - & 73.47 & 39 & - & - & - & 40.69 & - & - & - & - & - & 27.20 \\
\multirow{-7}{*}{InternVL3.5-8B} & Sliding Window & 2.29 & 2.33 & 4.99 & 5.49 & 2.58 & 66.33 & 36 & 15.98 & 2.68 & 38.30 & 42.67 & 17.02 & - & - & - & - & 24.45 \\
\bottomrule
\end{tabular}}
\end{table}

\subsection{Data Engine and Annotation}
UHR-Micro contains 1,212 curated images from established remote sensing datasets, including FAIR1Mv2 \citep{sun2022fair1m}, DOTAv2 \citep{xia2018dota}, SODA-A \citep{miri2025soda}, and xView \citep{lam2018xview}. To ensure extreme scale disparities, almost all selected images exceed 4,000 pixels along the long edge. Built on this UHR foundation, we design a three-tier hybrid data engine to balance scale and annotation quality. First, foundational tasks driven by coordinate or category mapping, such as \textit{GD}, \textit{OC}, and spatial relations, are programmatically generated from dense ground-truth annotations via geometric rules. Second, tasks requiring descriptive semantic alignment, including \textit{BG} and \textit{PDR}, are generated by a frontier VLM and then cross-validated by human annotators. Third, tasks involving deep reasoning, such as \textit{CG}, \textit{CRC}, and \textit{MCR}, are led by domain experts with model assistance and rigorous expert review. This pipeline scales the annotation process while preserving objective evaluation. All generated instructions are subjected to manual verification or rule-based consistency checks before inclusion. Detailed rules, prompts, and protocols are provided in Appendix~\ref{app:dataset_details}.

\subsection{Analysis}
UHR-Micro includes 11,072 deterministic evaluation instructions distributed across the 16 tasks. The instruction pool is partitioned into development, validation, and test splits. In our experiments, validation is used only for prompt-format checking and inference-protocol selection, while test is reserved for final benchmarking. Tab.~\ref{tab_datasets} compares UHR-Micro with existing vision-language datasets. UHR-Micro uniquely couples UHR imagery with extreme micro-targets: although the average long-edge resolution exceeds 4,000 pixels, target sizes and their relative proportions remain exceptionally small. This severe spatial disparity is designed to isolate micro-level visual extraction and reduce reliance on linguistic priors.

\section{Findings and Baseline}
To systematically address the challenges posed by the UHR-Micro benchmark, we first conduct a pilot study to analyze the behavior of representative VLMs under different visual perception strategies. The insights derived from this analysis directly motivate the design of our active perception method.

\subsection{Pilot Study and Empirical Findings}
We evaluated two representative models, Qwen3-VL-8B~\citep{bai2025qwen3} and InternVL3.5-8B~\citep{wang2025internvl3}, on the UHR-Micro validation split. To examine the impact of spatial resolution and localization strategies, we implement five comparative settings: 1) \textbf{Native:} Direct input of the original UHR image without external resizing or cropping; 2) \textbf{Resize:} Down-sampling the global image to a 1024-pixel long edge; 3) \textbf{Query Crop:} For tasks with explicit coordinate prompts, directly cropping the designated region; 4) \textbf{Oracle GT-Crop:} Cropping regions centered around the ground-truth targets at varying window sizes; and 5) \textbf{Sliding Window:} An exhaustive traversal of the global image using fixed-size patches.

\begin{figure}[t]
  \centering
  \includegraphics[width=\linewidth]{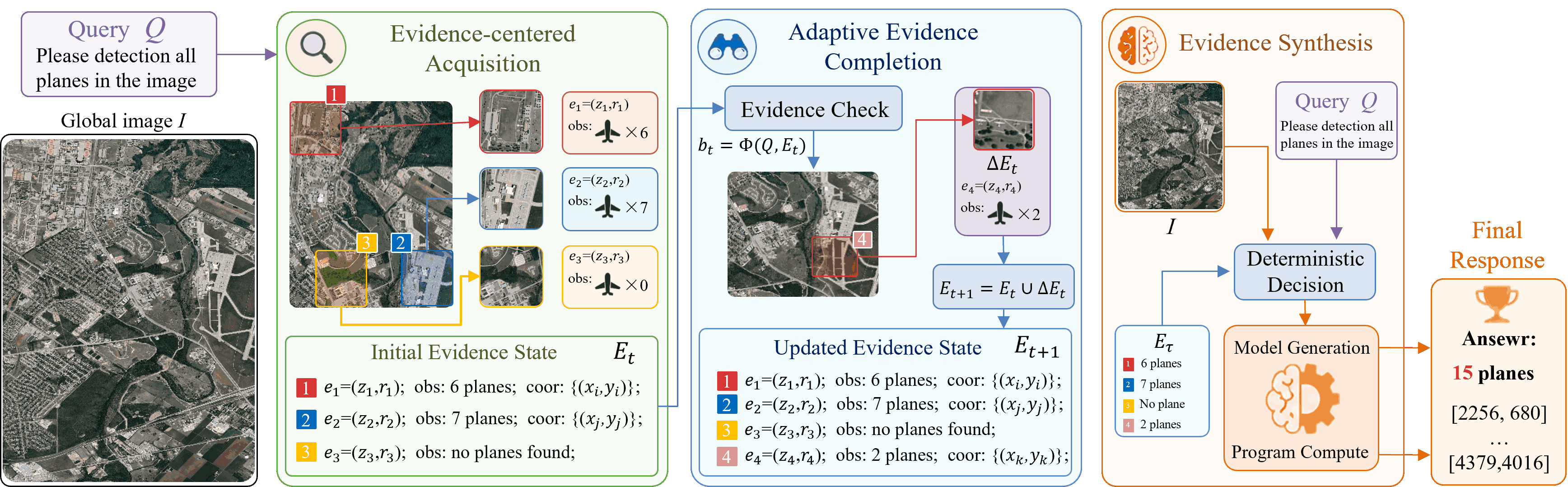}
\caption{
Overview of MAP. MAP acquires localized native-resolution observations and records each observation with its source region in the original image. These spatially grounded observations are combined with the global context to produce the final prediction. When the collected evidence is insufficient, MAP selectively acquires additional observations to complete the evidence state.
}
  \label{fig3}
\end{figure}

Summarized in Tab.~\ref{tab_val}, the quantitative results reveal three critical findings regarding micro-target perception:

\textbf{Finding 1: Large image size is not the sole bottleneck.} Performance generally decreases in the Resize-1024 setting compared to the Native setting, with the degradation especially pronounced for Qwen3-VL-8B~\citep{bai2025qwen3}. This suggests that the difficulty cannot be attributed simply to large image dimensions: reducing the global image size sacrifices critical micro-target detail without eliminating the underlying scale disparity.

\textbf{Finding 2: Localized evidence substantially improves micro-target recognition.} Both the Query Crop and Oracle GT-Crop strategies yield substantial gains over the Native setting, indicating that poor native performance is not solely caused by a lack of fine-grained recognition ability. Instead, much of the difficulty lies in finding the task-relevant evidence within the massive image canvas. Once localized, original-resolution crops are provided, VLMs can better exploit their latent recognition capability.

\textbf{Finding 3: The prohibitive cost of blind search.} While the Sliding Window strategy passively guarantees native-resolution coverage and improves metric performance in some tasks, it incurs substantial computational overhead. Specifically, this exhaustive strategy requires an average of 13 model calls per question, drastically inflating the computational cost for each image. Such redundancy makes exhaustive sliding-window search costly for practical UHR deployment.

Driven by these findings, we introduce MAP, a training-free reference method that investigates whether explicit evidence acquisition can alleviate the resolution illusion. MAP enables evidence-based inference by constructing and progressively refining a spatially grounded evidence state from native-resolution observations.

\subsection{Micro-evidence Active Perception}

As illustrated in Fig.~\ref{fig3}, given a UHR image $I$ and query $Q$, MAP acquires localized native-resolution observations and registers them to the source image. Each observation forms an evidence entry $e_i=(z_i,r_i)$, where $z_i$ is the parsed local observation and $r_i$ its source-image region. A complete-scene observation $g$ is retained to provide global context.

\textbf{Evidence-centered Acquisition.}
MAP first determines a compact set of question-conditioned support regions and constructs the initial evidence state as
$$
\mathcal{S}=\Pi(I,Q),\qquad
E_0=\left\{\Gamma\!\left(O(C(I,s),Q),s\right)\mid s\in\mathcal{S}\right\},
$$
where $\Pi$ identifies candidate support regions, $C$ extracts native-resolution views, $O$ obtains local observations, and $\Gamma$ registers them to the source-image frame.

\textbf{Targeted Evidence Completion.}
The initial evidence may leave some answer-relevant visual support missing or insufficiently resolved. MAP therefore iteratively identifies a region requiring further inspection and updates the evidence state as
$$
b_t=\Phi(Q,E_t),\qquad
E_{t+1}=E_t\cup
\left\{\Gamma\!\left(O(C(I,b_t),Q),b_t\right)\right\}.
$$
Here, $E_t$ denotes the evidence accumulated up to iteration $t$, $\Phi$ identifies missing or insufficient visual support from $E_t$ given the query, and $b_t$ denotes the corresponding source-image region selected for further observation.
This update selectively supplements missing evidence under a bounded observation budget rather than expanding visual coverage uniformly. The resulting evidence state $E_T$, together with the complete-scene observation $g$, is then used for final prediction. Details are provided in Appendix~\ref{app:map_details}.

\begin{table}[t]
\caption{
Performance on the UHR-Micro test split for representative models and MAP variants. Scores are reported as percentages. Category-level Avg. columns average the tasks within each category, while the final Avg. column averages all 16 task scores. The best and second-best results in each column are marked in bold and underlined, respectively.
}
\label{tab_test}
\centering
\resizebox{\textwidth}{!}{%
\begin{tabular}{@{}l|ccccc|c|cccc|c|cccc|c|ccc|c|c@{}}
\toprule
\multirow{2}{*}{Model}
& \multicolumn{6}{c|}{Grounding}
& \multicolumn{5}{c|}{Fine-grained Understanding}
& \multicolumn{5}{c|}{Counting}
& \multicolumn{4}{c|}{Spatial Reasoning}
& \multirow{2}{*}{Avg.} \\
\cmidrule(lr){2-7}
\cmidrule(lr){8-12}
\cmidrule(lr){13-17}
\cmidrule(lr){18-21}
& GD & RD & BG & CG & MCR & Avg.
& OC & FGR & RS & CS & Avg.
& GC & RC & CC & CRC & Avg.
& DrR & DsR & PDR & Avg.
& \\
\midrule
\multicolumn{21}{@{}l}{\textit{Open-source General-purpose VLMs}} \\
\midrule
Qwen2-VL-7B                  & 0.21 & 0.05 & 1.16 & 0.27 & 0.37 & 0.41 & 55.33 & 28.57 & 1.03 & 0.78 & 21.43 & 23.06 & 11.67 & 27.83 & 14.62 & 19.30 & 52.50 & 37.74 & 51.74 & 47.33 & 19.34 \\
Qwen2.5-VL-7B                & 1.75 & 0.00 & 7.07 & 3.38 & 1.80 & 2.80 & 58.59 & 37.24 & 6.99 & 2.05 & 26.22 & 26.44 & 18.41 & 39.30 & 19.61 & 25.94 & 63.12 & 50.31 & 67.44 & 60.29 & 25.96 \\
Qwen3-VL-8B                   & 5.79 & 0.04 & 19.59 & 10.37 & 4.42 & 8.04 & 64.26 & 36.22 & 19.63 & 5.85 & 31.49 & 20.37 & 12.08 & 15.96 & 18.25 & 16.66 & 68.12 & 70.44 & 75.00 & 71.19 & 26.55 \\
Qwen3-VL-32B                  & 4.02 & 0.53 & 16.92 & 10.43 & 6.31 & 7.64 & 65.12 & 34.69 & 18.67 & 7.19 & 31.42 & 20.70 & 12.75 & 13.67 & 15.67 & 15.70 & 76.88 & 76.10 & 82.56 & 78.51 & 28.77 \\
Qwen3-VL-235B-A22B            & \underline{10.19} & 2.09 & \textbf{28.72} & \textbf{18.21} & \textbf{11.08} & \underline{14.06} & \textbf{69.42} & 37.24 & \underline{27.92} & 7.71 & \textbf{35.57} & 26.64 & 14.32 & 16.12 & 17.79 & 18.72 & \underline{87.50} & 75.47 & \underline{84.88} & \underline{82.62} & \underline{33.30} \\ 
InternVL3.5-8B           & 4.03 & 0.70 & 8.87 & 4.01 & 3.51 & 4.23 & 58.59 & 32.65 & 18.11 & 2.64 & 28.00 & 35.33 & 25.42 & 23.88 & 22.59 & 26.80 & 45.00 & 37.74 & 47.67 & 43.47 & 22.51 \\
InternVL3.5-38B          & 9.14 & 2.09 & 20.49 & 8.49 & 10.29 & 10.10 & 66.15 & 38.78 & \textbf{28.25} & 4.61 & 34.45 & \underline{42.05} & 30.34 & 22.56 & \underline{26.01} & 30.24 & 53.12 & 41.51 & 47.09 & 47.24 & 28.69 \\
\midrule
\multicolumn{21}{@{}l}{\textit{Closed-source Frontier VLMs}} \\
\midrule
GPT-5.5                    & 0.51 & 1.08 & 2.75 & 0.90 & 0.46 & 1.14 & 39.69 & 31.63 & 3.00 & 2.99 & 19.33 & 37.12 & 10.40 & 26.09 & 25.45 & 24.77 & 81.88 & \underline{81.76} & 81.98 & 81.87 & 26.19 \\ 
Gemini-3.1-Pro & 1.76 & 0.16 & 6.71 & 9.66 & 2.14 & 4.09 & 54.98 & 33.16 & 4.13 & 1.55 & 23.46 & 36.67 & 20.96 & 31.91 & 25.89 & 28.86 & \textbf{96.25} & \textbf{84.28} & \textbf{95.93} & \textbf{92.15} & 30.67 \\ 
Claude Sonnet 5       & 0.21 & 0.63 & 0.93 & 0.49 & 0.16 & 0.48 & 39.18 & 30.61 & 0.69 & 1.42 & 17.98 & 14.13 & 12.35 & 15.20 & 19.53 & 15.30 & 85.00 & 40.25 & 60.47 & 61.91 & 16.67 \\
\midrule
\multicolumn{21}{@{}l}{\textit{General-domain Zoom-based Methods}} \\
\midrule
DeepEyes                 & 0.43 & 0.02 & 1.80 & 0.67 & 0.65 & 0.71 & 59.97 & 33.16 & 1.85 & 1.27 & 24.06 & 30.67 & 25.72 & 40.03 & 19.97 & 29.10 & 62.50 & 43.40 & 69.19 & 58.36 & 25.27 \\
HiDe     & 0.00 & 0.00 & 2.14 & 1.75 & 0.00 & 0.78 & 59.62 & 30.10 & 2.20 & 0.63 & 23.14 & 40.95 & 20.20 & 40.84 & 23.09 & 31.27 & 85.62 & 41.51 & 66.86 & 64.66 & 25.97 \\
ZoomEye  & 0.00 & 0.00 & 0.58 & 0.10 & 0.00 & 0.14 & 57.39 & 34.69 & 0.98 & 1.14 & 23.55 & 22.92 & 17.49 & \underline{44.87} & 23.85 & 27.28 & 23.75 & 32.70 & 42.44 & 32.96 & 18.94 \\
\midrule
\multicolumn{21}{@{}l}{\textit{Remote-sensing-oriented VLMs and Geospatial Agents}} \\
\midrule
GeoChat                & 0.09 & 0.00 & 0.22 & 0.01 & 0.26 & 0.12 & 45.70 & 29.59 & 0.05 & 0.11 & 18.87 & 25.87 & 18.23 & \textbf{46.88} & 19.96 & 27.74 & 27.50 & 22.01 & 20.93 & 23.48 & 16.09 \\
GeoLLaVa-8k               & 0.00 & 0.00 & 0.00 & 0.00 & 0.00 & 0.00 & 33.16 & 28.57 & 0.01 & 0.00 & 15.44 & 0.00 & 0.00 & 0.00 & 0.00 & 0.00 & 30.00 & 22.64 & 31.40 & 28.01 & 9.11 \\
GeoEyes                    & 0.36 & 0.00 & 1.99 & 1.44 & 0.00 & 0.76 & 54.30 & 32.14 & 2.43 & 0.37 & 22.31 & 28.22 & 22.44 & 25.36 & 20.07 & 24.02 & 57.50 & 40.88 & 64.53 & 54.30 & 22.00 \\
ZoomSearch                  & 4.96 & 0.05 & 6.60 & 7.18 & 2.13 & 4.18 & 64.78 & 34.69 & 6.16 & 3.97 & 27.40 & 25.68 & 11.42 & 15.39 & 17.07 & 17.39 & 58.13 & 47.80 & 58.14 & 54.69 & 22.76 \\
\midrule
\multicolumn{21}{@{}l}{\textit{MAP Variants}} \\
\midrule 
\makecell[l]{MAP (InternVL3.5-8B)} & 7.11 & \textbf{24.13} & 16.11 & 10.11 & \underline{10.80} & 13.65 & 62.71 & \textbf{41.84} & 19.85 & \underline{9.49} & 33.47 & \textbf{43.19} & \textbf{57.81} & 30.86 & \textbf{30.30} & \textbf{40.54} & 44.38 & 38.36 & 53.49 & 45.41 & 30.68 \\
\makecell[l]{MAP (Qwen3-VL-8B)}    & \textbf{15.70} & \underline{13.17} & \underline{22.79} & \underline{13.48} & 9.51 & \textbf{14.93} & \underline{68.38} & \underline{39.29} & 20.71 & \textbf{12.36} & \underline{35.18} & 36.87 & \underline{45.09} & 24.21 & 22.81 & \underline{32.25} & 79.38 & 67.92 & 76.74 & 74.68 & \textbf{35.68} \\
\bottomrule
\end{tabular}}
\end{table}

\section{Experiment}
\subsection{Experimental Setup}
Following the held-out split described above, all final results are reported on the test split of UHR-Micro. The validation split is used only to fix inference-time protocol details before test evaluation, and no model weights are updated. We evaluate five groups of models. The first group contains open-source general-purpose VLMs, including Qwen2-VL-7B~\citep{wang2024qwen2}, Qwen2.5-VL-7B~\citep{hui2024qwen25}, Qwen3-VL-8B~\citep{bai2025qwen3}, Qwen3-VL-32B, Qwen3-VL-235B-A22B, InternVL3.5-8B~\citep{wang2025internvl3}, and InternVL3.5-38B. The second group contains closed-source frontier VLMs, including GPT-5.5~\citep{singh2025openai}, Gemini-3.1-Pro~\citep{team2023gemini}, and Claude Sonnet 5~\citep{bai2022claude}. The third group contains general-domain zoom-based methods, including DeepEyes~\citep{zheng2026deepeyes}, HiDe~\citep{liu2026hide} and ZoomEye~\citep{shen2025zoomeye}.
The fourth group contains remote-sensing-oriented VLMs and geospatial agents, including GeoChat~\citep{kuckreja2024geochat}, GeoLLaVA-8K~\citep{wang2025geollava}, GeoEyes~\citep{wang2026geoeyes}, and ZoomSearch~\citep{zhou2025zoomsearch}. The fifth group contains MAP variants built on top of the Qwen3-VL-8B and InternVL3.5-8B baselines. All models are evaluated on the same image-query pairs, with coordinate outputs converted to absolute pixels for scoring; we use a 1000-base protocol by default and model-native protocols when specified, with all protocol choices fixed before test evaluation. For segmentation tasks, we use a standardized box-to-mask protocol: models first predict bounding boxes, which are then converted to absolute pixels and used as SAM2~\citep{ravi2024sam2} box prompts for mask generation. MAP inspects at most eight $1024\times1024$ observation windows per question, and acquires at most four additional $640\times640$ views when the recorded observations are insufficient to answer the question. Additional implementation details are provided in Appendix~\ref{app:map_details}.

\subsection{Evaluation Metrics}
UHR-Micro contains heterogeneous answer formats, including categorical choices, counts, box localization, multi-region detection, and segmentation. We define all task scores in the range of $[0,1]$ and report them as percentages for readability. The overall score is computed as the macro-average over tasks. Empty, invalid, or unparsable predictions are assigned zero.

Answers are read only from the model's final output; surrounding explanations, reasoning traces, and echoes of the question are never searched for candidate answers. For categorical tasks, we use exact accuracy after normalizing model outputs to canonical labels. For counting tasks, we use a bounded relative counting score rather than exact match, so that errors are penalized according to their magnitude relative to the ground-truth count. For single-target localization, we use a continuous localization score that combines box overlap with center proximity and scale agreement, and that applies to both horizontal and oriented boxes. For multi-region detection and retrieval, predicted and ground-truth boxes are matched by one-to-one optimal assignment, and the score is a soft F1 over the matched localization scores. For segmentation tasks, we use an analogous continuous formulation based on mask overlap, centroid proximity, and area consistency. Detailed metric definitions are provided in Appendix~\ref{app:metrics}.

\subsection{Main Results}
Tab.~\ref{tab_test} reports the main results on the UHR-Micro test split. Overall performance remains limited across model families, with grounding emerging as the most consistent bottleneck. For instance, Gemini-3.1-Pro~\citep{team2023gemini} reaches 92.15 on spatial reasoning, yet its grounding average remains only 4.09. Such a gap would be obscured in conventional scene-level benchmarks, highlighting the diagnostic value of UHR-Micro.

Scaling and domain specialization only partially alleviate this bottleneck. Within the Qwen3-VL family, the overall score rises from 26.55 for Qwen3-VL-8B~\citep{bai2025qwen3} to 33.30 for Qwen3-VL-235B-A22B~\citep{bai2025qwen3}. Closed-source frontier models show similarly uneven behavior: Gemini-3.1-Pro~\citep{team2023gemini}, GPT-5.5~\citep{singh2025openai} and Claude Sonnet 5~\citep{bai2022claude} show similarly uneven performance across task families. Remote-sensing-oriented models and geospatial agents remain limited, suggesting that, among evaluated models, domain specialization alone does not yield strong micro-evidence localization among the evaluated models.

MAP directly targets this weakness. Built on Qwen3-VL-8B~\citep{bai2025qwen3}, it improves the overall score from 26.55 to 35.68, surpassing the strongest non-MAP model. The largest category-level gains occur in grounding and counting, consistent with MAP's focus on acquiring localized task-relevant evidence. The same trend holds for InternVL3.5-8B~\citep{wang2025internvl3}, where MAP improves the overall score by 8.17 points.

Overall, current VLMs are not uniformly weak in visual reasoning, but suffer from a specific bottleneck in accessing and using micro-evidence. Higher input resolution, larger capacity, and remote-sensing specialization provide limited gains, whereas MAP shifts UHR reasoning from image-centered perception toward evidence-centered reasoning, and it delivers consistent gains.

\begin{figure}[t]
\centering
\noindent

\begin{minipage}[t]{0.49\textwidth}
\vspace{0pt}
\centering

\includegraphics[width=\linewidth]{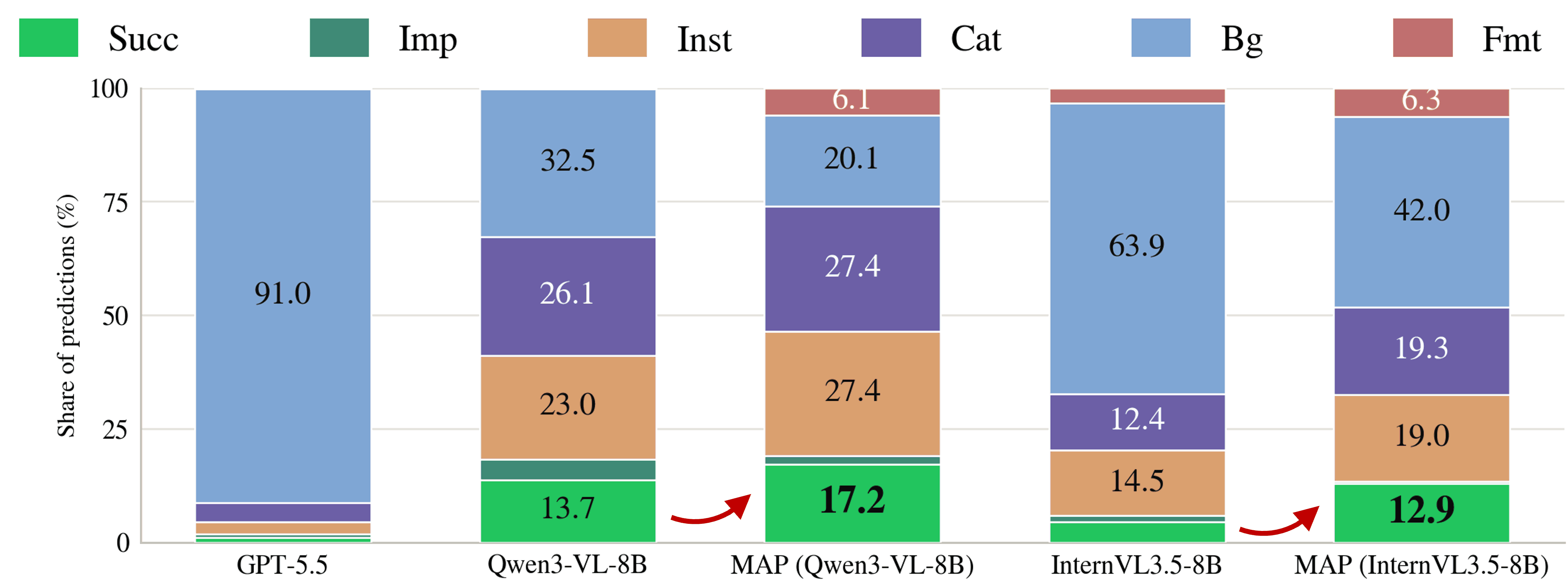}

\vspace{1mm}
\captionof{figure}{
Error diagnosis on the \textit{CG} task.
Fmt: format failure; Imp: imprecise localization;
Inst: instance mismatch; Cat: category mismatch;
Bg: background; Succ: success.
}
\label{fig:error_diag}

\end{minipage}
\hfill
\begin{minipage}[t]{0.49\textwidth}
\vspace*{-2mm}
\centering

\captionof{table}{
Component ablation of MAP on the UHR-Micro validation split.
}
\label{tab_map_components}

\vspace{1mm}

\resizebox{\linewidth}{!}{
\begin{tabular}{@{}lccccc@{}}
\toprule
Config. & Grd. & Und. & Cnt. & Rel. & Avg. \\
\midrule
\multicolumn{6}{@{}l}{\textit{Qwen3-VL-8B}} \\
\midrule
Baseline
& 5.74 & 31.40 & 15.91 & 75.00 & 27.68 \\
+ Acquisition
& 10.87 & 32.88 & 31.23 & 79.00 & 34.24 \\
+ Completion
& 13.25 & 32.88 & 35.06 & 79.67 & \textbf{36.06} \\
\midrule
\multicolumn{6}{@{}l}{\textit{InternVL3.5-8B}} \\
\midrule
Baseline
& 3.02 & 24.24 & 33.43 & 41.00 & 23.05 \\
+ Acquisition
& 7.20 & 32.93 & 34.42 & 45.33 & 27.59 \\
+ Completion
& 10.09 & 32.93 & 38.51 & 45.33 & \textbf{29.51} \\
\bottomrule
\end{tabular}
}

\end{minipage}

\vspace{-2mm}
\end{figure}

\subsection{Diagnostic Analysis}
To understand where micro-evidence grounding breaks down, we diagnose the predictions on the CG task of the UHR-Micro test split. We analyze GPT-5.5~\citep{singh2025openai}, Qwen3-VL-8B~\citep{bai2025qwen3}, InternVL3.5-8B~\citep{wang2025internvl3} and the corresponding MAP variants. Each prediction is assigned to one category by rules that only compare it with the ground-truth box and the annotated objects of the same image, so the analysis is reproducible and requires no manual labeling. A prediction counts as successful (Succ.) when its localization score against the ground-truth reaches 0.3; otherwise we ask what went wrong: \textbf{Fmt.} violates the output contract; for a valid box, \textbf{Imp.} finds the correct instance but localizes it imprecisely, \textbf{Inst.} hits a different instance of the same category, \textbf{Cat.} hits another category, and \textbf{Bg.} lands on no annotated object at all.

As shown in Fig.~\ref{fig:error_diag}, background predictions constitute a major failure mode on this task, particularly for the closed-source frontier VLM. This pattern is consistent with the extreme scale disparity in UHR-Micro: task-relevant targets occupy only a tiny fraction of the image, making unguided localization particularly prone to background errors. MAP substantially reduces these background failures on both backbones. Among the remaining MAP errors, predictions increasingly reach an annotated target but select the wrong instance or category. This shift suggests that MAP moves the bottleneck from finding object-level evidence toward selecting the correct task-relevant evidence among competing candidates.

\subsection{Ablation Study}

Tab.~\ref{tab_map_components} evaluates MAP through component ablation. Evidence-centered Acquisition consistently improves both backbones, showing that question-guided native-resolution evidence collection is effective for UHR perception. Targeted Evidence Completion provides further gains, with the largest improvements appearing in grounding and counting. The consistent gains across both backbones suggest that MAP is not specific to a single VLM architecture. Details are provided in Appendix~\ref{app:map_ablation_detail}.

\section{Conclusion}
In this work, we presented UHR-Micro, a diagnostic benchmark for micro-scale perception and reasoning in ultra-high-resolution Earth observation imagery. It contains 11,072 instructions across 1,212 UHR images and 16 tasks, with task-relevant targets occupying extremely small regions. Our evaluation shows that higher input resolution, model scale, and remote-sensing specialization alone do not ensure reliable access to micro-evidence. We further introduced MAP, an evidence-centered active perception method that adaptively acquires and supplements question-relevant evidence, improving average performance by 8.65 points across two backbone VLMs. These results highlight that reliable micro-level reasoning requires not only more pixels, but effective access to task-critical evidence. Remaining errors in localization and fine-grained disambiguation point to important directions for future UHR VLM research.

\bibliography{iclr2027_conference}
\bibliographystyle{iclr2027_conference}

\clearpage
\appendix
\section{Appendix}

\phantomsection

\begingroup
\small
\setlength{\parindent}{0pt}
\setlength{\parskip}{2pt}

\newcommand{\appcontentsline}[2]{%
\noindent\hyperref[#1]{#2}\dotfill\hyperref[#1]{\pageref*{#1}}\par}

\newcommand{\appcontentsubline}[2]{%
\noindent\hspace*{1.5em}\hyperref[#1]{#2}\dotfill\hyperref[#1]{\pageref*{#1}}\par}

\appcontentsline{app:dataset_details}{\textbf{Appendix A. Dataset Construction Details}}
\appcontentsubline{app:source_filtering}{A.1 Source Datasets and Image Filtering}
\appcontentsubline{app:task_schema}{A.2 Task Taxonomy and Instruction Schema}
\appcontentsubline{app:split_statistics}{A.3 Split Construction and Data Statistics}

\medskip

\appcontentsline{app:data_engine}{\textbf{Appendix B. Data Engine and Annotation Protocol}}
\appcontentsubline{app:programmatic_generation}{B.1 Programmatic Task Derivation}
\appcontentsubline{app:vlm_generation}{B.2 VLM-assisted Semantic Task Generation}
\appcontentsubline{app:expert_annotation}{B.3 Expert-led Annotation for Complex Tasks}
\appcontentsubline{app:quality_control}{B.4 Human Verification and Quality Control}

\medskip

\appcontentsline{app:metrics}{\textbf{Appendix C. Evaluation Protocol and Metrics}}
\appcontentsubline{app:output_normalization}{C.1 Output Format Normalization}
\appcontentsubline{app:bbox_metric}{C.2 Bounding Box Evaluation}
\appcontentsubline{app:mask_metric}{C.3 Segmentation Evaluation}
\appcontentsubline{app:count_option_metric}{C.4 Counting and Option Evaluation}
\appcontentsubline{app:score_aggregation}{C.5 Overall Score Aggregation}
\appcontentsubline{app:metric_sensitivity}{C.6 Sensitivity to Localization Metric Hyperparameters}

\medskip

\appcontentsline{app:map_details}{\textbf{Appendix D. MAP Implementation Details}}
\appcontentsubline{app:map_algorithm}{D.1 End-to-End Inference Procedure}
\appcontentsubline{app:roi_discovery}{D.2 Evidence-centered Acquisition}
\appcontentsubline{app:evidence_state}{D.3 Spatially Grounded Evidence State}
\appcontentsubline{app:local_inspection}{D.4 Targeted Evidence Completion}
\appcontentsubline{app:evidence_synthesis}{D.5 Evidence Aggregation and Answer Generation}
\appcontentsubline{app:map_prompts}{D.6 Prompt Interfaces and Output Contracts}
\appcontentsubline{app:map_cost}{D.7 Inference Budget and Hyperparameters}

\medskip

\appcontentsline{app:additional_analyses}{\textbf{Appendix E. Additional Analyses and Protocol Details}}
\appcontentsubline{app:bgcg_factor_analysis}{E.1 Sensitivity to Scale}
\appcontentsubline{app:coord_protocols}{E.2 Coordinate Protocols}
\appcontentsubline{app:map_ablation_detail}{E.3 Detailed MAP Ablations}

\medskip

\appcontentsline{app:qualitative_examples}{\textbf{Appendix F. Qualitative Examples of UHR-Micro}}

\medskip

\appcontentsline{app:datasheets}{\textbf{Appendix G. Datasheets}}

\endgroup

\subsection{Dataset Construction Details}
\label{app:dataset_details}

This section provides additional details on the construction of UHR-Micro, including source data selection, task schemas, and split construction.

\subsubsection{Source Datasets and Image Filtering}
\label{app:source_filtering}

\paragraph{DOTAv2.}
DOTAv2~\citep{xia2018dota} is a large-scale benchmark for object detection in aerial images, containing high-resolution scenes collected from diverse sensors and platforms. Its images exhibit substantial variations in object scale, orientation, density, and scene layout, making it well suited for constructing UHR tasks that require both global scene awareness and precise localization. DOTAv2 provides expert annotations in the form of arbitrary quadrilateral boxes over common aerial object categories, which allows us to derive coordinate-based grounding, detection, counting, and spatial-relation tasks. In UHR-Micro, we retain DOTAv2 images with 4K-scale long-edge resolution and sufficient target density and spatial diversity for deterministic micro-target evaluation.

\paragraph{FAIR1Mv2.}
FAIR1Mv2~\citep{sun2022fair1m} is designed for fine-grained object recognition in high-resolution remote sensing imagery. Compared with generic aerial detection datasets, FAIR1M provides richer category granularity, including fine-grained subcategories for objects such as aircraft, ships, vehicles, sports fields, and road-related targets. Its oriented bounding-box annotations and fine-grained labels are particularly useful for constructing tasks that require semantic discrimination among visually similar micro-targets, such as object classification and fine-grained recognition. We select UHR scenes from FAIR1Mv2 with 4K-scale long-edge resolution and reliable fine-grained annotations, enabling UHR-Micro to test whether VLMs can distinguish subtle target types under severe scale disparity.

\paragraph{SODA-A.}
SODA-A~\citep{miri2025soda} is the aerial subset of the SODA benchmark, which was introduced to support large-scale small object detection. It focuses on aerial scenes with dense small objects and provides oriented box annotations across multiple object categories. This makes SODA-A especially relevant to UHR-Micro, where the central challenge is not merely object recognition but the discovery of extremely small task-relevant evidence within a large image canvas. We use SODA-A images to enrich micro-target density and to construct tasks involving dense enumeration, local discrimination, and crowded-scene grounding. Images are filtered to ensure UHR resolution, valid annotations, and deterministic target-answer correspondence.

\paragraph{xView.}
xView~\citep{lam2018xview} is a large-scale overhead imagery dataset collected from WorldView-3 satellite imagery at 0.3m ground sample distance. It contains diverse geographic scenes and a broad taxonomy of object categories, with dense bounding-box annotations over small, rare, and fine-grained object types. The dataset is valuable for UHR-Micro because it introduces real-world satellite imagery with complex backgrounds, heterogeneous object distributions, and challenging small-object instances. We retain xView scenes that meet the UHR resolution criterion and contain sufficient annotated objects for constructing grounding, counting, and spatial reasoning instructions. All selected annotations are converted into the unified UHR-Micro schema before task generation.

\paragraph{Unified filtering and standardization.}
Across all source datasets, we apply a common filtering and standardization pipeline. We primarily retain UHR images with a long edge of at least 4,000 pixels, while selectively including suitable images above 3,500 pixels. Images are required to contain multiple valid annotated targets and support deterministic task construction. Corrupted images, incomplete annotations, duplicated samples, and cases whose answers cannot be deterministically verified are removed.

\paragraph{Retained source statistics.}
Tab.~\ref{tab:source_statistics} summarizes the source-wise composition and resolution statistics of UHR-Micro after filtering.

\begin{table}[t]
\centering
\caption{
Source-wise image statistics of UHR-Micro after filtering. Average resolutions are rounded to the nearest hundred pixels.
}
\label{tab:source_statistics}
\resizebox{0.8\textwidth}{!}{
\begin{tabular}{lcccc}
\toprule
Source Dataset & Images & Avg. Resolution & Avg. Pixels & Max. Resolution \\
\midrule
DOTAv2~\citep{xia2018dota}    & 145 & $4800 \times 4200$ & 24.10 MP & $29200 \times 27620$ \\
FAIR1Mv2~\citep{sun2022fair1m}    & 77  & $5700 \times 5800$ & 34.42 MP & $10000 \times 9472$ \\
SODA-A~\citep{miri2025soda}     & 863 & $4800 \times 2800$ & 13.26 MP & $4762 \times 4800$ \\
xView~\citep{lam2018xview}      & 127 & $4300 \times 3100$ & 13.28 MP & $4994 \times 3187$ \\
\midrule
Total                          & 1212 & $4800 \times 3200$ & 15.90 MP & $29200 \times 27620$ \\
\bottomrule
\end{tabular}
}
\end{table}

\subsubsection{Task Taxonomy and Instruction Schema}
\label{app:task_schema}

UHR-Micro defines a unified instruction schema over 16 tasks, covering four complementary dimensions of micro-target perception: grounding, fine-grained understanding, counting, and spatial reasoning. Each instruction is paired with a deterministic target representation, allowing model responses to be parsed in a consistent and objective manner. Since remote sensing annotations naturally involve both horizontal and oriented bounding-box representations, UHR-Micro does not impose a single geometric convention across all localization tasks. Instead, the required answer representation is explicitly specified in each instruction.

\paragraph{Geometric answer representations.}
For localization-oriented tasks, including \textit{GD}, \textit{RD}, \textit{BG}, \textit{CG}, and \textit{MCR}, targets may be represented as either horizontal bounding boxes or oriented bounding boxes. A horizontal bounding box is encoded as $[x_1,y_1,x_2,y_2]$, while an oriented bounding box is encoded by four polygon vertices as $[x_1,y_1,x_2,y_2,x_3,y_3,x_4,y_4]$. Multi-target tasks require a sequence of such geometric primitives. In the released annotations, all target coordinates are stored as absolute pixel coordinates on the original image canvas. This design preserves the spatial precision required for UHR evaluation while remaining compatible with heterogeneous source annotations; model-facing prompts may use normalized coordinate conventions, while model predictions are converted back to absolute pixel coordinates before scoring.

\begin{table}[t]
\centering
\caption{
Instruction schema of UHR-Micro. The table summarizes the input condition and deterministic answer representation associated with each task.
}
\label{tab:task_schema}
\resizebox{\textwidth}{!}{
\begin{tabular}{llll}
\toprule
Dimension & Task & Input Condition & Answer Representation \\
\midrule
\multirow{5}{*}{Grounding}
& Global Detection (GD)
& Full-image query
& Multiple horizontal or oriented bounding boxes \\
& Regional Detection (RD)
& Region-constrained query
& Multiple horizontal or oriented bounding boxes \\
& Basic Grounding (BG)
& Referring expression over the full image
& Single horizontal or oriented bounding box \\
& Complex Grounding (CG)
& multiple spatial or relational constraints
& Single horizontal or oriented bounding box \\
& Multi-Condition Retrieval (MCR)
& Multi-attribute and spatial-condition query
& Multiple horizontal or oriented bounding boxes \\
\midrule
\multirow{4}{*}{Fine-grained Understanding}
& Object Classification (OC)
& Indicated target in the full image
& Multiple-choice option \\
& Fine-Grained Recognition (FGR)
& Indicated target instance
& Multiple-choice option \\
& Referring Segmentation (RS)
& Referring expression over the full image
& COCO-style compressed RLE mask \\
& Component Segmentation (CS)
& Component-level expression
& COCO-style compressed RLE mask \\
\midrule
\multirow{4}{*}{Counting}
& Global Counting (GC)
& Full-image query
& Integer answer \\
& Regional Counting (RC)
& Region-constrained query
& Integer answer \\
& Conditional Counting (CC)
& Attribute- or region-conditioned query
& Integer answer \\
& Cross-Region Compare (CRC)
& Pair of regions or region descriptions
& Integer answer \\
\midrule
\multirow{3}{*}{Spatial Reasoning}
& Directional Relationship (DrR)
& Marked object pair
& Multiple-choice option \\
& Distance Relationship (DsR)
& Reference object and candidate objects
& Multiple-choice option \\
& Pattern Distribution Recognition (PDR)
& Full image with distribution options
& Multiple-choice option \\
\bottomrule
\end{tabular}
}
\end{table}

\paragraph{Region and visual-reference conventions.}
Region-conditioned tasks use heterogeneous but deterministic forms of spatial reference. In the annotation files, \textit{RD} and \textit{RC} primarily use rectangular regions stored in absolute pixel coordinates, which are converted to the model-facing coordinate protocol during benchmarking; \textit{CRC} may involve either coordinate-defined regions or language-described regions. \textit{CC} often uses natural-language regions of interest. For relational tasks, \textit{DrR} and \textit{DsR} identify reference objects and candidates through rendered visual markers. \textit{PDR} operates on the full image, where the answer is selected from candidate distribution descriptions.

\paragraph{Segmentation representations.}
For both \textit{RS} and \textit{CS}, the released target is represented as a COCO-style compressed RLE mask, which can be decoded into a binary mask using the corresponding image size. \textit{RS} focuses on object-level segmentation from referring expressions, whereas \textit{CS} targets component-level segmentation from part descriptions.

\subsubsection{Split Construction and Data Statistics}
\label{app:split_statistics}

UHR-Micro is released with development, validation, and test partitions. The development partition provides diverse task instances for method exploration and prompt design. The validation partition is used for selecting inference-time configurations, such as crop size, region budget, and inference strategies. The test partition is used for reporting final benchmark results. Unless otherwise specified, all results in the main paper are reported on the test partition. No model weights are updated using the development or validation partitions.

Each instruction is paired with a deterministic target representation and evaluated under its corresponding task protocol. This design allows UHR-Micro to probe different perceptual skills, spatial regions, object instances, and reasoning conditions in UHR imagery while maintaining objective and reproducible evaluation.

\begin{table}[t]
\centering
\caption{
Task-wise instruction statistics of UHR-Micro.
}
\label{tab:task_split_statistics}
\resizebox{0.45\textwidth}{!}{
\begin{tabular}{lrrrr}
\toprule
Task & Development & Validation & Test & Total \\
\midrule
GD  & 647 & 97  & 556 & 1300 \\
RD  & 276 & 100 & 296 & 672  \\
BG  & 628 & 100 & 568 & 1296 \\
CG  & 370 & 100 & 379 & 849  \\
MCR & 173 & 100 & 204 & 477  \\
OC  & 672 & 98  & 582 & 1352 \\
FGR & 163 & 100 & 196 & 459  \\
RS  & 115 & 100 & 149 & 364  \\
CS  & 98  & 100 & 108 & 306  \\
GC  & 296 & 100 & 304 & 700  \\
RC  & 569 & 100 & 544 & 1213 \\
CC  & 169 & 100 & 202 & 471  \\
CRC & 209 & 90  & 170 & 469  \\
DrR & 113 & 100 & 160 & 373  \\
DsR & 112 & 100 & 159 & 371  \\
PDR & 128 & 100 & 172 & 400  \\
\midrule
Total & 4738 & 1585 & 4749 & 11072 \\
\bottomrule
\end{tabular}
}
\end{table}

As shown in Tab.~\ref{tab:task_split_statistics}, UHR-Micro contains 11,072 deterministic instructions across 16 tasks, including 4,738 development instructions, 1,585 validation instructions, and 4,749 test instructions. The test split is image-disjoint from the development and validation splits, with no source image overlapping those used for tuning, avoiding image-level overlap between test and development/validation data.

\subsection{Data Engine and Annotation Protocol}
\label{app:data_engine}

This section details the three-tier hybrid data engine introduced in the main paper. UHR-Micro is built upon dense annotations from source remote sensing datasets, combining programmatic derivation, VLM-assisted semantic generation, and expert-guided construction to balance scalability, semantic diversity, and reliable supervision. The three tiers correspond to increasing task complexity: deterministic perception tasks are derived from geometric and categorical annotations; semantic alignment tasks leverage frontier VLMs with validation; complex reasoning tasks are constructed through expert-guided annotation. All generated samples undergo automatic consistency checks and expert review before release, with answers constrained to verifiable formats, including bounding boxes, segmentation masks, exact counts, and discrete choices.

\subsubsection{Programmatic Task Derivation}
\label{app:programmatic_generation}

The first tier constructs perception tasks whose answers can be derived from dense source annotations. Object categories, coordinates, instance geometries, and spatial relations provide deterministic supervision for tasks including detection, recognition, counting, and spatial reasoning.

\paragraph{Localization and recognition.}
Localization tasks are generated by grouping annotated instances according to image, category, and spatial scope. Global detection considers the full image, while regional detection focuses on specified areas containing valid targets. Ground-truth locations are retained in the original absolute-pixel coordinate system and represented as horizontal or oriented bounding boxes. Recognition tasks associate instances with annotation labels, with related distractors introduced for fine-grained classification when applicable.

\paragraph{Counting and spatial reasoning.}
Counting tasks derive exact numerical answers through annotation-based enumeration, including Global Counting (GC) and Regional Counting (RC). Spatial-relation tasks are generated from object geometries and evaluate directional or distance relationships. Candidate samples are filtered and reviewed to remove ambiguous configurations.

\subsubsection{VLM-assisted Semantic Task Generation}
\label{app:vlm_generation}

The second tier targets tasks requiring semantic alignment beyond direct annotation mapping. Frontier VLMs assist in generating referring expressions, descriptions, and candidate options, while final answers remain determined by source annotations or curated target sets. Generated language is filtered and reviewed to ensure semantic correctness, target uniqueness, and consistency with visual evidence.

\paragraph{Semantic grounding and distribution recognition.}
Selected grounding tasks use VLM-generated referring expressions to identify annotated targets based on appearance, spatial, or contextual cues. Distribution recognition tasks use VLMs to describe object layouts and generate candidate answers. All language outputs are treated as auxiliary annotations rather than ground truth and are verified before inclusion.

\subsubsection{Expert-led Annotation for Complex Tasks}
\label{app:expert_annotation}

The third tier addresses tasks requiring multi-step reasoning, complex constraints, or region-level semantic interpretation. Experts define reasoning conditions and refine model-assisted annotations to ensure unique and visually grounded answers. This tier covers complex grounding, multi-condition retrieval, conditional counting, cross-region comparison, and mask-level understanding.

\paragraph{Complex reasoning and numerical tasks.}
Experts construct instructions involving spatial relations, object interactions, regional layouts, and multiple constraints. For retrieval and conditional counting, valid target sets and conditions are explicitly defined to ensure deterministic answers. Cross-region comparison samples are verified for valid regions and reliable numerical comparisons, while ambiguous cases are removed.

\paragraph{Mask-level annotation.}
Referring and component segmentation tasks provide mask-level targets in COCO-style compressed RLE format. Object-level masks are associated with textual descriptions, while component segmentation focuses on fine-grained object structures. Masks are reviewed to ensure correspondence with the intended target and meaningfulness at UHR scale.

\subsubsection{Human Verification and Quality Control}
\label{app:quality_control}

All generated samples undergo automatic validation before release, with human review applied according to task complexity. The quality-control process checks deterministic targets, valid output formats, and visual grounding while minimizing subjective interpretation. Automatic checks verify structural correctness, including coordinate validity, format consistency, duplicate removal, option uniqueness, and numerical consistency. Programmatically generated samples are extensively spot-checked, while human annotators focus on semantic correctness, ambiguity removal, and visual evidence verification for language-heavy and reasoning-intensive tasks.

Samples are further filtered based on task-specific criteria. Localization and segmentation outputs are checked for valid geometric representations, counting and comparison tasks are verified against annotation-derived targets, and language-based instructions are reviewed for target uniqueness, condition completeness, and semantic faithfulness. Samples with invalid formats, inconsistent annotations, ambiguous descriptions, duplicate answers, or insufficient visual evidence are removed.

This hybrid validation strategy preserves the scalability of automatic generation while maintaining reliable evaluation targets for micro-level perception in UHR imagery.

\subsection{Evaluation Protocol and Metrics}
\label{app:metrics}

UHR-Micro adopts a deterministic evaluation protocol with format-specific metrics to accommodate heterogeneous task outputs. All scores are normalized to $[0,1]$ and reported as percentages. Invalid, empty, or unparsable predictions receive zero scores. This design enables consistent comparison across task families while preserving the semantics of each output format.

\subsubsection{Output Format Normalization}
\label{app:output_normalization}

Before evaluation, model responses are converted into canonical formats specified by each task instruction. Since VLMs may generate free-form or partially structured outputs, a lightweight parser is applied to the model's designated final output to convert valid predictions into canonical formats. The normalized outputs are then transformed into the required representation for metric computation.

For localization and segmentation tasks, predicted coordinates are parsed and converted to the original image coordinate system before geometric scoring or mask generation. Predictions with invalid formats, inconsistent dimensions, or degenerate structures are rejected. For counting tasks, the parser extracts a unique integer answer; ambiguous or invalid numerical responses are treated as incorrect. For discrete-choice tasks, equivalent textual variants are mapped to canonical option labels, while conflicting or missing choices are considered invalid.

\subsubsection{Bounding Box Evaluation}
\label{app:bbox_metric}

Bounding-box evaluation is used for localization tasks, including detection, grounding, and retrieval. Since micro-targets occupy only a small fraction of UHR images, strict IoU thresholds can be overly sensitive to small coordinate shifts, while center-distance metrics ignore target extent. We therefore define a continuous localization score combining overlap, center alignment, and scale consistency. All predictions are converted to native image coordinates before evaluation.

\paragraph{Single-target scoring.}
Let $g$ and $p$ denote the ground-truth and predicted boxes. Horizontal and oriented boxes are converted to polygon regions for overlap computation. The localization score is defined as

\[
S_{\mathrm{box}}(g,p)
=
\mathrm{IoU}(g,p)
+
0.5(1-\mathrm{IoU}(g,p))
C(g,p)\Lambda(g,p),
\]

where

\[
C(g,p)=
\max(0,1-\frac{d(g,p)}{2\sigma_g})
\]

measures normalized center proximity, and

\[
\Lambda(g,p)=
\sqrt{\frac{\min(A_g,A_p)}
{\max(A_g,A_p)}}
\]

measures scale consistency based on box areas. Here $d(g,p)$ is the centroid distance and $\sigma_g$ is the diagonal length of the ground-truth enclosing box. The score is bounded in $[0,1]$, providing partial credit for predictions that are spatially close but not perfectly aligned.

For single-target grounding tasks, the predicted box is directly scored against the unique target. Invalid outputs receive zero scores. When multiple boxes are returned, only the first valid prediction is evaluated according to the predefined output contract.

\paragraph{Multi-target scoring.}
For multi-target tasks, let $P=\{p_1,\dots,p_m\}$ and $G=\{g_1,\dots,g_n\}$. We compute pairwise scores $Q_{ij}=S_{\mathrm{box}}(g_i,p_j)$ and perform one-to-one Hungarian matching over valid pairs with $Q_{ij}\geq0.3$. Let

\[
T=\sum_{(i,j)\in\mathcal{M}}Q_{ij}
\]

be the total matched score. The final localization score is defined as

\[
F_{\mathrm{soft}}
=
\frac{2T}{|P|+|G|}.
\]

This formulation jointly accounts for localization accuracy, missed targets, and false positives without requiring confidence-ranked predictions. It is therefore suitable for free-form VLM outputs, which typically lack calibrated detection scores.

\subsubsection{Segmentation Evaluation}
\label{app:mask_metric}

Segmentation tasks are evaluated using mask-level targets. Under our protocol, predicted regions are converted into masks using a fixed SAM2-based postprocessor. Similar to box localization, we adopt a continuous mask score to reduce the brittleness of hard IoU thresholds for extremely small targets.

Let $m$ and $\hat m$ denote the ground-truth and predicted masks. The score is defined as

\[
S_{\mathrm{mask}}(m,\hat m)
=
\mathrm{IoU}_{\mathrm{mask}}(m,\hat m)
+
0.5(1-\mathrm{IoU}_{\mathrm{mask}}(m,\hat m))
C(m,\hat m)A_{\mathrm{ratio}}(m,\hat m),
\]

where $C(m,\hat m)$ measures normalized centroid proximity following the box metric, and

\[
A_{\mathrm{ratio}}(m,\hat m)
=
\frac{\min(|m|,|\hat m|)}
{\max(|m|,|\hat m|)}
\]

measures foreground-area consistency. The score is bounded in $[0,1]$ and provides partial credit for masks that are spatially close and have similar extent. Invalid predictions, failed mask generation, or undecodable outputs receive zero scores.

\subsubsection{Counting and Option Evaluation}
\label{app:count_option_metric}

Counting tasks use a bounded relative-error score. Given the ground-truth and predicted counts $c$ and $\hat c$,

\[
M_{\mathrm{cnt}}(c,\hat c)=
\begin{cases}
\max(0,1-\frac{|c-\hat c|}{c}), & c>0,\\
\mathbf{1}[\hat c=0], & c=0.
\end{cases}
\]

This metric assigns full credit to exact counts and decreases with relative error, while invalid numerical outputs receive zero.

Discrete-option tasks are evaluated using exact accuracy:

\[
M_{\mathrm{opt}}(y,\hat y)=\mathbf{1}[\hat y=y].
\]

Since these tasks have closed answer spaces, exact matching provides deterministic evaluation after option normalization.

\subsubsection{Overall Score Aggregation}
\label{app:score_aggregation}

All sample-level scores are normalized to $[0,1]$ according to their answer format. For each task $t$, the task score is computed as

\[
S_t=\frac{1}{N_t}\sum_{i=1}^{N_t}s_{t,i},
\]

where invalid or missing predictions receive zero scores. The final UHR-Micro score is the macro-average over all 16 tasks:

\[
S_{\mathrm{overall}}
=
\frac{1}{16}\sum_{t=1}^{16}S_t .
\]

Category-level scores are obtained by averaging task scores within each evaluation category. Macro-averaging prevents tasks with more samples from dominating the benchmark and ensures that each task contributes equally to the overall score. Reported results multiply normalized scores by 100, with rounding applied only for display.

\subsubsection{Robustness of Localization Evaluation}
\label{app:metric_sensitivity}

We examine whether the main grounding conclusions depend on the hyperparameters of our continuous localization metric. We consider two components: the auxiliary-term weight $w$ in single-target localization and the matching threshold $\tau$ used for multi-target assignment. In all sensitivity experiments, model predictions are kept fixed and only the evaluation hyperparameter is varied.

\paragraph{Sensitivity to the auxiliary-term weight.}
For single-target localization, the score is

$$
S_{\mathrm{box}}(g,p)
=
\mathrm{IoU}(g,p)
+
w(1-\mathrm{IoU}(g,p))
C(g,p)\Lambda(g,p),
$$

with $w=0.5$ fixed in the main experiments. We vary $w$ from $0$ to $1.0$ and evaluate BG and CG. Since larger $w$ assigns greater partial credit to predictions with better center and scale alignment, absolute scores increase with $w$ by construction. We therefore focus on the stability of relative model comparisons and MAP gains.

\begin{table}[t]
\centering
\caption{
Sensitivity to the auxiliary-term weight $w$ on the UHR-Micro test split. The default value used in the main experiments is $w=0.5$. Predictions are fixed across settings and only the evaluation weight is varied. Scores are percentages.
}
\label{tab:metric_weight_sensitivity}
\resizebox{0.75\textwidth}{!}{
\begin{tabular}{llccccc}
\toprule
Task & Model & $w=0$ & $w=0.3$ & $w=0.5$ & $w=0.7$ & $w=1.0$ \\
\midrule
\multirow{6}{*}{BG}
& GPT-5.5                  & 0.98  & 2.05  & 2.75  & 3.46  & 4.53 \\
& Gemini-3.1-Pro           & 4.24  & 5.72  & 6.71  & 7.70  & 9.18 \\
& Qwen3-VL-8B              & 13.50 & 17.16 & 19.59 & 22.03 & 25.69 \\
& InternVL3.5-8B           & 4.92  & 7.29  & 8.87  & 10.45 & 12.83 \\
& MAP (Qwen3-VL-8B)        & 17.47 & 20.67 & 22.79 & 24.92 & 28.12 \\
& MAP (InternVL3.5-8B)     & 11.45 & 14.25 & 16.11 & 17.97 & 20.77 \\
\midrule
\multirow{6}{*}{CG}
& GPT-5.5                  & 0.29  & 0.65  & 0.90  & 1.14  & 1.51 \\
& Gemini-3.1-Pro           & 5.63  & 8.05  & 9.66  & 11.27 & 13.69 \\
& Qwen3-VL-8B              & 5.91  & 8.58  & 10.37 & 12.16 & 14.83 \\
& InternVL3.5-8B           & 1.45  & 2.99  & 4.01  & 5.04  & 6.58 \\
& MAP (Qwen3-VL-8B)        & 9.42  & 11.86 & 13.48 & 15.11 & 17.54 \\
& MAP (InternVL3.5-8B)     & 6.50  & 8.67  & 10.11 & 11.55 & 13.71 \\
\bottomrule
\end{tabular}
}
\end{table}

As shown in Tab.~\ref{tab:metric_weight_sensitivity}, the relative ordering of the evaluated systems remains unchanged across all tested values of $w$ on both BG and CG. More importantly, both MAP variants consistently outperform their corresponding direct-inference backbones throughout the full range. Although absolute scores vary with $w$, the relative comparisons and the improvement brought by MAP remain stable.

\paragraph{Sensitivity to the multi-target matching threshold.}
For multi-target localization, pairwise localization scores are matched by one-to-one Hungarian assignment, with pairs retained only when $Q_{ij}\geq\tau$. The main experiments use $\tau=0.3$. We additionally evaluate $\tau\in{0.5,0.7}$ on GD, RD, and MCR. Increasing $\tau$ imposes a stricter requirement on matched predictions and therefore decreases absolute scores by construction; we again focus on whether the comparative conclusions remain stable.

\begin{table}[t]
\centering
\caption{
Sensitivity to the multi-target matching threshold $\tau$ on the UHR-Micro test split. The default value used in the main experiments is $\tau=0.3$. Predictions are fixed across settings and only the matching threshold is varied. Scores are percentages.
}
\label{tab:metric_threshold_sensitivity}
\resizebox{0.6\textwidth}{!}{
\begin{tabular}{llccc}
\toprule
Task & Model & $\tau=0.3$ & $\tau=0.5$ & $\tau=0.7$ \\
\midrule
\multirow{6}{*}{GD}
& GPT-5.5                  & 0.51  & 0.25  & 0.15 \\
& Gemini-3.1-Pro           & 1.76  & 1.43  & 0.91 \\
& Qwen3-VL-8B              & 5.79  & 4.31  & 2.74 \\
& InternVL3.5-8B           & 4.03  & 2.94  & 1.56 \\
& MAP (Qwen3-VL-8B)        & 15.70 & 14.38 & 11.37 \\
& MAP (InternVL3.5-8B)     & 7.11  & 6.38  & 4.16 \\
\midrule
\multirow{6}{*}{RD}
& GPT-5.5                  & 1.08  & 0.45  & 0.05 \\
& Gemini-3.1-Pro           & 0.16  & 0.07  & 0.01 \\
& Qwen3-VL-8B              & 0.04  & 0.00  & 0.00 \\
& InternVL3.5-8B           & 0.70  & 0.30  & 0.03 \\
& MAP (Qwen3-VL-8B)        & 13.17 & 12.77 & 9.67 \\
& MAP (InternVL3.5-8B)     & 24.13 & 21.06 & 11.80 \\
\midrule
\multirow{6}{*}{MCR}
& GPT-5.5                  & 0.46  & 0.25  & 0.00 \\
& Gemini-3.1-Pro           & 2.14  & 1.59  & 1.07 \\
& Qwen3-VL-8B              & 4.42  & 3.71  & 2.68 \\
& InternVL3.5-8B           & 3.51  & 2.51  & 1.62 \\
& MAP (Qwen3-VL-8B)        & 9.51  & 9.23  & 7.83 \\
& MAP (InternVL3.5-8B)     & 10.80 & 10.07 & 7.14 \\
\bottomrule
\end{tabular}
}
\end{table}

Tab.~\ref{tab:metric_threshold_sensitivity} shows that stricter matching thresholds lower absolute scores, as expected, but preserve the main comparative trends. On GD and RD, the relative ordering of the evaluated systems remains unchanged across all tested thresholds. On MCR, the two MAP variants exchange their relative order at $\tau=0.7$, while both remain ahead of the corresponding direct-inference backbones and the other evaluated models. Across all three multi-target tasks, MAP therefore retains its advantage under substantially stricter matching criteria.

\paragraph{Overlap-only localization evaluation.}
We further test whether the grounding results persist when the auxiliary center- and scale-consistency terms are removed entirely. Predictions are kept fixed and re-evaluated using polygon IoU alone. For the single-target tasks BG and CG, we report Success@IoU$\eta$, i.e., the percentage of samples whose predicted geometry overlaps the ground-truth geometry by at least $\eta$; missing or invalid predictions are counted as failures. For the multi-target tasks GD, RD, and MCR, predictions are matched one-to-one to ground-truth instances using IoU-only greedy matching, and we report macro precision, recall, and F1 across samples. We evaluate both $\eta=0.25$ and $\eta=0.50$.

\begin{table}[t]
\centering
\caption{
Strict overlap-only evaluation on single-target grounding. Success@IoU$\eta$ is the percentage of samples whose predicted geometry achieves polygon IoU of at least $\eta$ with the ground truth. Scores are percentages.
}
\label{tab:strict_iou_single}
\resizebox{0.78\textwidth}{!}{
\begin{tabular}{lcccc}
\toprule
Model & BG@0.25 & BG@0.50 & CG@0.25 & CG@0.50 \\
\midrule
GPT-5.5                         & 1.76  & 0.35  & 0.26  & 0.00 \\
Gemini-3.1-Pro-Preview         & 6.51  & 3.70  & 11.35 & 3.69 \\
Qwen3-VL-8B                    & 21.66 & 10.74 & 9.50  & 3.69 \\
InternVL3.5-8B                 & 8.28  & 3.87  & 2.38  & 0.53 \\
MAP (Qwen3-VL-8B)              & \textbf{25.70} & \textbf{17.43} & \textbf{14.78} & \textbf{8.97} \\
MAP (InternVL3.5-8B)           & 19.54 & 12.85 & 10.55 & 6.60 \\
\bottomrule
\end{tabular}}
\end{table}

\begin{table}[t]
\centering
\caption{
Strict overlap-only evaluation on multi-target grounding. Predictions are matched using IoU only. We report macro precision (P), recall (R), and F1 across samples at IoU thresholds 0.25 and 0.50. Scores are percentages.
}
\label{tab:strict_iou_multi}
\resizebox{\textwidth}{!}{
\begin{tabular}{llcccccc}
\toprule
& & \multicolumn{3}{c}{IoU$\geq0.25$} & \multicolumn{3}{c}{IoU$\geq0.50$} \\
\cmidrule(lr){3-5}\cmidrule(lr){6-8}
Task & Model & P & R & F1 & P & R & F1 \\
\midrule
\multirow{6}{*}{GD}
& GPT-5.5                       & 0.49  & 0.58  & 0.51  & 0.18  & 0.18  & 0.18 \\
& Gemini-3.1-Pro-Preview       & 2.51  & 2.37  & 2.20  & 1.07  & 1.23  & 1.07 \\
& Qwen3-VL-8B                  & 9.42  & 6.69  & 7.00  & 4.45  & 3.53  & 3.52 \\
& InternVL3.5-8B               & 5.68  & 5.91  & 4.80  & 2.38  & 2.58  & 2.09 \\
& MAP (Qwen3-VL-8B)            & \textbf{25.22} & \textbf{19.05} & \textbf{19.50} & \textbf{18.57} & \textbf{13.80} & \textbf{14.24} \\
& MAP (InternVL3.5-8B)         & 11.67 & 9.24  & 9.28  & 7.41  & 5.92  & 5.91 \\
\midrule
\multirow{6}{*}{RD}
& GPT-5.5                       & 1.32  & 0.84  & 0.88  & 0.13  & 0.07  & 0.09 \\
& Gemini-3.1-Pro-Preview       & 0.16  & 0.12  & 0.13  & 0.02  & 0.01  & 0.02 \\
& Qwen3-VL-8B                  & 0.01  & 0.04  & 0.01  & 0.00  & 0.01  & 0.00 \\
& InternVL3.5-8B               & 0.85  & 1.25  & 0.86  & 0.03  & 0.08  & 0.04 \\
& MAP (Qwen3-VL-8B)            & 19.44 & 17.66 & 17.14 & 14.25 & 13.10 & 12.97 \\
& MAP (InternVL3.5-8B) & \textbf{34.06} & \textbf{40.87} & \textbf{32.26} & \textbf{17.12} & \textbf{20.85} & \textbf{16.48} \\
\midrule
\multirow{6}{*}{MCR}
& GPT-5.5                       & 0.51  & 0.42  & 0.45  & 0.00  & 0.00  & 0.00 \\
& Gemini-3.1-Pro-Preview       & 2.72  & 2.08  & 2.29  & 1.72  & 1.16  & 1.35 \\
& Qwen3-VL-8B                  & 6.47  & 5.42  & 5.43  & 3.79  & 2.97  & 3.10 \\
& InternVL3.5-8B               & 3.87  & 4.92  & 3.98  & 2.05  & 2.28  & 2.06 \\
& MAP (Qwen3-VL-8B)            & 12.81 & 14.76 & 11.43 & \textbf{10.73} & 12.89 & \textbf{9.87} \\
& MAP (InternVL3.5-8B)         & \textbf{15.20} & \textbf{22.02} & \textbf{14.66} & 9.18 & \textbf{13.89} & 8.78 \\
\bottomrule
\end{tabular}}
\end{table}

The overlap-only results reinforce the conclusions obtained with the continuous localization metric. On both BG and CG, MAP improves Success@IoU over its corresponding direct-inference backbone at both thresholds. The improvement remains substantial even at IoU$\geq0.50$: MAP increases BG success from 10.74 to 17.43 for Qwen3-VL-8B and from 3.87 to 12.85 for InternVL3.5-8B, while CG improves from 3.69 to 8.97 and from 0.53 to 6.60, respectively.

The same pattern holds for multi-target grounding. Across GD, RD, and MCR, both MAP variants achieve higher macro F1 than their corresponding direct backbones under both IoU thresholds. At IoU$\geq0.50$, for example, MAP improves F1 from 3.52 to 14.24 on GD and from 0.00 to 12.97 on RD with Qwen3-VL-8B, while the corresponding InternVL3.5-8B improvements are 2.09 to 5.91 and 0.04 to 5.44. These gains persist when localization credit is determined solely by geometric overlap, showing that MAP's grounding improvements are not an artifact of the auxiliary center- or scale-consistency terms.

Overall, the three analyses provide complementary evidence for the robustness of our localization evaluation. Varying the auxiliary weight $w$ and matching threshold $\tau$ preserves the main comparative trends of the continuous metric. More importantly, replacing the proposed score with strict IoU-only evaluation preserves MAP's advantage over its corresponding direct-inference backbone across all five grounding tasks and both overlap thresholds. Thus, the reported improvements do not depend on a particular choice of metric hyperparameters or on partial credit from center and scale proximity.

\subsection{MAP Implementation Details}
\label{app:map_details}

This section specifies the implementation of MAP following the formulation in Sec.~4.2. Given an image--query pair $(I,Q)$, MAP constructs an initial evidence state $E_0$ from question-conditioned native-resolution observations, iteratively supplements missing evidence when necessary, and produces the final prediction from the completed evidence state $E_T$ together with a complete-scene observation $g$.

\subsubsection{End-to-End Inference Procedure}
\label{app:map_algorithm}

MAP proceeds in three stages. First, Evidence-centered Acquisition determines a compact set of spatial supports $\mathcal{S}$ and converts observations from these regions into the initial evidence state $E_0$. Second, Targeted Evidence Completion uses the current state $E_t$ to identify additional regions whose visual support is missing or insufficient, and incorporates newly acquired observations into $E_{t+1}$. Finally, the accumulated evidence is either aggregated by a predefined schema-level rule or passed, together with the global context, to a single composition step for final prediction. The following sections specify each stage.

\subsubsection{Evidence-centered Acquisition}
\label{app:roi_discovery}

The support operator $\Pi$ determines the spatial supports used for initial local observation,

$$
\mathcal{S}=\Pi(I,Q).
$$

In our implementation, model-proposed supports are represented as normalized points $p_i=(u_i,v_i)$ in the $[0,1000]$ image frame and mapped to source-image coordinates by

$$
(x_i,y_i)=\left(\frac{u_iW}{1000},\frac{v_iH}{1000}\right),
$$

where $W$ and $H$ are the source-image dimensions. Each point defines a $1024\times1024$ native-resolution observation window centered at $(x_i,y_i)$; windows near image boundaries are shifted to remain within the source image without resizing. Spatial supports already specified by the query are preserved directly.

For each support $s\in\mathcal{S}$, MAP extracts a native-resolution view $C(I,s)$, obtains a local observation $O(C(I,s),Q)$, and registers the result to the source-image frame through $\Gamma$. The initial evidence state is therefore

$$
E_0=
\left\{
\Gamma\!\left(O(C(I,s),Q),s\right)
\,\middle|\,
s\in\mathcal{S}
\right\}.
$$

A complete-scene observation $g$ is obtained once from the full image and retained as complementary global context.

\subsubsection{Spatially Grounded Evidence State}
\label{app:evidence_state}

Each registered observation is represented as

$$
e_i=(z_i,r_i),
$$

where $z_i$ is the parsed structured observation and $r_i=[x_1,y_1,x_2,y_2]$ records its support region in the source-image frame. Depending on the required output schema, $z_i$ is represented as a scalar, discrete option, or source-registered geometry.

Geometric predictions made within a local view are transformed back to the common source-image coordinate frame before being added to the evidence state. Consequently, observations acquired from different regions or different acquisition stages can be accumulated in the same representation,

$$
E_t=\{e_i\}_{i=1}^{N_t}.
$$

\subsubsection{Targeted Evidence Completion}
\label{app:local_inspection}

The initial state $E_0$ may not cover all visual support required by the query. MAP therefore uses a completion operator $\Phi$ to identify a source-image region that requires further inspection,

$$
b_t=\Phi(Q,E_t).
$$

The selected region $b_t$ is converted into additional native-resolution views, which are processed through the same observation and registration pipeline. The evidence state is then updated as

$$
E_{t+1}
=
E_t
\cup
\left\{
\Gamma\!\left(O(C(I,b_t),Q),b_t\right)
\right\}.
$$

For larger support regions, $b_t$ is deterministically covered by native-resolution windows of up to $640\times640$ pixels. Additional observations are bounded to at most four views per query. Completion continues only while further visual support is identified and the observation budget remains available.

\subsubsection{Evidence Aggregation and Answer Generation}
\label{app:evidence_synthesis}

After evidence acquisition and completion, MAP produces the final answer from $E_T$. Structured evidence is aggregated deterministically whenever the required output can be directly derived from the registered observations. Our implementation uses three schema-level cases:

\begin{itemize}[leftmargin=*]
\item \textbf{Scalar composition:} numerical relations supported by registered scalar observations are computed directly.
\item \textbf{Direct structured output:} when a complete supported observation already provides the required structured answer, it is returned directly.
\item \textbf{Geometry-set composition:} registered geometry candidates are merged in the source-image frame and deduplicated using non-maximum suppression.
\end{itemize}

For all remaining cases, a single composition step receives the query $Q$, complete-scene observation $g$, and accumulated evidence state $E_T$ to produce the final response under the required output format. Segmentation outputs are obtained from the final predicted support using the fixed SAM2 postprocessor described in Sec.~\ref{app:mask_metric}.

\subsubsection{Prompt Interfaces and Output Contracts}
\label{app:map_prompts}

MAP uses a small set of prompt interfaces corresponding to the main inference stages. Table~\ref{tab:map_prompt_interfaces} summarizes their inputs and structured outputs. Across stages, prompts constrain the model to produce intermediate observations rather than unrestricted answers, while task variation is primarily expressed through the query context and required output schema.

\begin{table}[t]
\centering
\caption{Prompt interfaces used by MAP.}
\label{tab:map_prompt_interfaces}
\resizebox{0.9\textwidth}{!}{
\begin{tabular}{lll}
\toprule
Stage & Input & Structured output \\
\midrule
Complete-scene observation
& Full image, query
& Provisional global observation \\
Support proposal
& Full image, query
& Normalized support points \\
Local observation
& Native-resolution view, localized query
& Schema-constrained local observation \\
Evidence completion
& Query, current evidence state
& Additional source-image support region \\
Final composition
& Query, global context, accumulated evidence
& Final task-formatted answer \\
\bottomrule
\end{tabular}}
\end{table}

\paragraph{Support proposal.}
The support proposer is instructed to return only a JSON list of distinct points in the source-image $[0,1000]$ coordinate frame,

$$
[[x_1,y_1],\ldots,[x_K,y_K]].
$$

These points are subsequently converted into native-resolution observation regions as described in Sec.~\ref{app:roi_discovery}.

\paragraph{Local observation.}
Local observations follow a shared observation interface that restricts reasoning to the visible source region and constrains the response to the required structured representation. The principal output contracts are summarized in Tab.~\ref{tab:map_output_contracts}.

\begin{table}[t]
\centering
\caption{Structured output contracts for local observations.}
\label{tab:map_output_contracts}
\begin{tabular}{ll}
\toprule
Output schema & Local observation \\
\midrule
Scalar & One non-negative integer \\
Option & One discrete option, or null if unsupported \\
Single geometry & One AABB4/QUAD8 geometry, or null \\
Geometry set & All visible AABB4/QUAD8 geometries, or null \\
\bottomrule
\end{tabular}
\end{table}

\paragraph{Evidence completion.}
The completion interface receives the query together with the current source-registered evidence and identifies at most one additional source-image region requiring visual inspection. The returned region is expressed in the same normalized source-image coordinate frame and is converted deterministically into the additional native-resolution observations used to update $E_t$.

\paragraph{Final composition.}
The final composition interface receives the query, complete-scene observation, accumulated evidence state, and required output contract. It produces a single final response in the benchmark format when deterministic aggregation does not already determine the answer.

The prompt templates and output constraints are fixed during evaluation; backbone-specific implementations preserve the same stage-level interfaces and structured output contracts.

\paragraph{Algorithmic summary.}
The MAP inference procedure can be summarized as follows:
\begin{enumerate}[leftmargin=*,nosep]
\item Obtain a complete-scene observation $g$ and initial supports $\mathcal{S}=\Pi(I,Q)$.
\item Construct
\(    E_0=
    \left\{
    \Gamma\!\left(O(C(I,s),Q),s\right)
    \,\middle|\,
    s\in\mathcal{S}
    \right\}.
   \)
\item Starting from $E_0$, identify additional support $b_t=\Phi(Q,E_t)$ when further visual evidence is required, and update
\(    E_{t+1}
    =
    E_t\cup
    \left\{
    \Gamma\!\left(O(C(I,b_t),Q),b_t\right)
    \right\}
   \)
until no further support is identified or the observation budget is reached.
\item Produce the final prediction from the completed evidence state $E_T$ and global context $g$ using the aggregation procedure in Sec.~\ref{app:evidence_synthesis}.
\end{enumerate}

\subsubsection{Inference Budget and Hyperparameters}
\label{app:map_cost}

MAP uses deterministic decoding throughout inference. Initial local observations use native-resolution windows of $1024\times1024$ pixels, with a compact support budget for localized queries and a larger budget for queries requiring spatially distributed evidence. Targeted Evidence Completion acquires at most four additional native-resolution views, each no larger than $640\times640$ pixels. Redundant initial observation regions are suppressed before local inference, and registered geometry candidates are deduplicated during final aggregation.

Table~\ref{tab:map_hyperparameters} summarizes the principal settings used in all experiments.

\begin{table}[t]
\centering
\caption{Main inference settings of MAP.}
\label{tab:map_hyperparameters}
\resizebox{0.55\textwidth}{!}{
\begin{tabular}{lc}
\toprule
Setting & Value \\
\midrule
Initial observation size & $1024\times1024$ \\
Localized support budget & 1--2 \\
Distributed support budget & 4--8 \\
Maximum additional views & 4 \\
Additional observation size & $\leq640\times640$ \\
Initial-region deduplication IoU & 0.2 \\
Geometry NMS IoU & 0.5 \\
Decoding temperature & 0.0 \\
Top-$p$ & 1.0 \\
\bottomrule
\end{tabular}}
\end{table}

For efficiency analysis, Calls/QA denotes the average number of backbone VLM invocations required for each query. The number of calls varies across samples because additional observations and final composition are invoked only when required.

\subsection{Additional Analyses and Protocol Details}
\label{app:additional_analyses}

This section provides additional analyses and implementation details beyond the main results. We first examine the sensitivity of localization performance to target scale, then summarize the coordinate protocols used across different model families, and finally provide detailed task-level ablations of MAP.

\subsubsection{Sensitivity to Scale}
\label{app:bgcg_factor_analysis}

We analyze how localization performance varies with target area on four fine-grained localization tasks: BG, CG, RS, and CS. Target area is measured using the ground-truth polygon area for box-based tasks and foreground pixel count for segmentation tasks. Tab.~\ref{tab:size_correlation} reports Pearson and Spearman correlations.

At the object level, target area remains an important factor for BG and RS. On BG, all systems show positive correlations, with the direct InternVL3.5-8B baseline exhibiting the strongest dependence ($r=0.348$). MAP reduces this correlation ($r=0.122$) while substantially improving grounding performance, suggesting that evidence-guided perception reduces the reliance on target scale. On RS, the size-performance relationship is stronger, with most open models showing consistent positive correlations, reflecting that larger masks provide more tolerance under pixel-level evaluation.

In contrast, target area has limited explanatory power for CG. Only direct baselines show moderate correlations, while MAP variants are nearly uncorrelated. This indicates that once grounding requires multi-hop relational constraints, difficulty is increasingly determined by factors beyond object extent, such as contextual and spatial reasoning requirements.

For CS, target area no longer separates performance across samples. All systems achieve low scores over the evaluated scale range, indicating that component-level targets remain challenging even when their sizes vary. These results suggest that current VLMs still struggle with fine-grained structural perception at the component level, which remains an important challenge for UHR image understanding.

Overall, target area primarily governs object-level localization tasks, while higher-level reasoning and component-level perception introduce additional bottlenecks beyond scale alone. MAP alleviates scale sensitivity in object-level grounding but does not yet fully resolve component-level perception.

\begin{table}[t]
\centering
\caption{
Correlation between target size and per-sample localization score on the test split. Target size is the ground-truth target area in pixels (box polygon area for BG/CG, mask area for RS/CS). The pooled columns merge the samples of all five systems within a task.
}
\label{tab:size_correlation}
\resizebox{0.75\textwidth}{!}{
\begin{tabular}{llrrrr}
\toprule
Task & Model & Pearson $r$ & Spearman $\rho$ & Pooled $r$ & Pooled $\rho$ \\
\midrule
\multirow{5}{*}{BG}
& Gemini-3.1-Pro~\citep{team2023gemini}  & 0.124 & 0.240 & \multirow{5}{*}{0.191} & \multirow{5}{*}{0.291} \\
& Qwen3-VL-8B~\citep{bai2025qwen3}  & 0.211 & 0.399 & & \\
& InternVL3.5-8B~\citep{wang2025internvl3}  & 0.348 & 0.390 & & \\
& MAP (Qwen3-VL-8B) & 0.204 & 0.324 & & \\
& MAP (InternVL3.5-8B) & 0.122 & 0.200 & & \\
\midrule
\multirow{5}{*}{CG}
& Gemini-3.1-Pro~\citep{team2023gemini}  & 0.035 & $-0.071$ & \multirow{5}{*}{0.059} & \multirow{5}{*}{0.013} \\
& Qwen3-VL-8B~\citep{bai2025qwen3}  & 0.150 & 0.036 & & \\
& InternVL3.5-8B~\citep{wang2025internvl3}  & 0.245 & 0.177 & & \\
& MAP (Qwen3-VL-8B) & 0.017 & $-0.055$ & & \\
& MAP (InternVL3.5-8B) & $-0.037$ & 0.022 & & \\
\midrule
\multirow{5}{*}{RS}
& Gemini-3.1-Pro~\citep{team2023gemini}  & $-0.028$ & 0.143 & \multirow{5}{*}{0.303} & \multirow{5}{*}{0.442} \\
& Qwen3-VL-8B~\citep{bai2025qwen3}  & 0.393 & 0.547 & & \\
& InternVL3.5-8B~\citep{wang2025internvl3}  & 0.411 & 0.561 & & \\
& MAP (Qwen3-VL-8B) & 0.376 & 0.438 & & \\
& MAP (InternVL3.5-8B) & 0.274 & 0.536 & & \\
\midrule
\multirow{5}{*}{CS}
& Gemini-3.1-Pro~\citep{team2023gemini}  & 0.016 & 0.168 & \multirow{5}{*}{0.105} & \multirow{5}{*}{0.211} \\
& Qwen3-VL-8B~\citep{bai2025qwen3}  & 0.096 & 0.328 & & \\
& InternVL3.5-8B~\citep{wang2025internvl3}  & 0.159 & 0.272 & & \\
& MAP (Qwen3-VL-8B) & 0.138 & 0.283 & & \\
& MAP (InternVL3.5-8B) & 0.141 & 0.172 & & \\
\bottomrule
\end{tabular}
}
\end{table}

\subsubsection{Coordinate Protocols}
\label{app:coord_protocols}

All coordinate-based tasks are evaluated in the source-image pixel frame. Predictions are first converted from their model-specific coordinate protocol into absolute source-image coordinates and are then compared with the annotations. This unified conversion ensures that models using different coordinate conventions are evaluated in the same spatial reference system.

Coordinate protocols are determined using a fixed procedure. When a model provides an explicit coordinate convention, we follow its native protocol. For models without a stable documented convention, we evaluate predefined candidate protocols on the validation split and freeze the selected protocol before test evaluation. This avoids ambiguity in coordinate interpretation while preventing protocol selection from depending on test results.

Most general-purpose VLMs, including Qwen and InternVL families, use normalized coordinates in the $[0,1000]$ frame. Remote-sensing models follow their checkpoint-specific conventions, including $[0,100]$, $[0,1]$, or absolute coordinates on the model-input canvas. Closed-source APIs are evaluated using a fixed source-image $[0,1000]$ contract. External zoom-based methods follow their released or frozen evaluation protocols, with necessary transformations applied to map predictions back to the source image. The complete protocol assignment is summarized in Tab.~\ref{tab:coord_protocols}.

\begin{table}[t]
\centering
\caption{
Coordinate protocols used by different model families.
All predictions are converted into absolute pixels in the original image coordinate system before evaluation.
}
\label{tab:coord_protocols}

\resizebox{\textwidth}{!}{
\begin{tabular}{llll}
\toprule
Model Family & Models & Coordinate Convention & Protocol Source \\
\midrule

\multirow{2}{*}{General-purpose VLMs}
& Qwen2-VL, Qwen2.5-VL, Qwen3-VL
& Normalized image coordinates ($[0,1000]$)
& Model-native convention \\
& InternVL3.5
& Normalized image coordinates ($[0,1000]$)
& Model-native convention \\

\midrule

General-purpose VLMs
& GPT-5.5, Gemini-3.1-Pro, Claude Sonnet 5
& Normalized image coordinates ($[0,1000]$)
& Fixed evaluation convention \\

\midrule

Zoom-based Visual Search Methods
& DeepEyes
& Absolute pixels on source image
& Released evaluation protocol \\

& HiDe, ZoomEye
& Absolute pixels on source image
& Validation-based calibration \\

\midrule

Remote Sensing VLMs
& GeoChat
& Normalized image coordinates ($[0,100]$)
& Model-native convention \\

& GeoLLaVA-8K
& Normalized image coordinates ($[0,1]$)
& Model-native convention \\

& GeoEyes
& Absolute pixels on model-input canvas
& Model-native convention \\

& ZoomSearch
& Normalized image coordinates ($[0,1000]$)
& Native composite coordinates with source-frame remapping \\

\midrule

MAP (Ours)
& MAP (InternVL3.5-8B), MAP (Qwen3-VL-8B)
& Normalized image coordinates ($[0,1000]$)
& MAP output convention \\

\bottomrule
\end{tabular}
}
\end{table}

\subsubsection{Detailed MAP Ablations}
\label{app:map_ablation_detail}

Table~\ref{tab_map_ablation_detail} provides a task-level ablation of MAP and compares it with Query Crop, Sliding Window, and representative zoom-based methods. Evidence-centered Acquisition yields substantial improvements across both backbones, while Targeted Evidence Completion further strengthens grounding and counting by selectively resolving missing evidence. Unlike Query Crop, MAP can discover relevant evidence when its location is not explicitly specified, and unlike Sliding Window, it avoids exhaustive visual coverage.

The Calls/QA results further highlight the efficiency of evidence-centered perception. MAP obtains most of its gains with a moderate number of VLM calls, while targeted completion introduces only a small additional cost. In contrast, exhaustive or iterative search requires substantially more calls without comparable performance gains. Notably, with the same Qwen3-VL-8B backbone, MAP substantially outperforms ZoomSearch while using fewer calls, demonstrating the advantage of targeted evidence acquisition over generic visual search.

\begin{table*}[t]
\centering
\caption{
Detailed ablation of MAP and comparison with Query Crop and Sliding Window on the UHR-Micro validation split.
}
\label{tab_map_ablation_detail}
\resizebox{\textwidth}{!}{%
\begin{tabular}{l|ccccc|cccc|cccc|ccc|c|c}
\toprule
& \multicolumn{5}{c|}{Grounding} & \multicolumn{4}{c|}{Fine-grained Understanding} & \multicolumn{4}{c|}{Counting} & \multicolumn{3}{c|}{Spatial Reasoning} & \\
\multirow{-2}{*}{Config.} & GD & RD & BG & CG & MCR & OC & FGR & RS & CS & GC & RC & CC & CRC & DrR & DsR & PDR & \multirow{-2}{*}{Avg.$^\dagger$} & \multirow{-2}{*}{Calls/QA} \\
\midrule
\multicolumn{19}{@{}l}{\textit{Qwen3-VL-8B}} \\
\midrule
Baseline                       & 2.10 & 0.47 & 16.33 & 5.55 & 4.22 & 76.53 & 29 & 13.77 & 6.29 & 23.40 & 12.87 & 12.80 & 14.56 & 75 & 76 & 74 & 27.68 & 1.00\\
+ Evidence-centered Acquisition & 7.60 & 7.99 & 23.80 & 10.25 & 4.71 & 67.35 & 35 & 15.51 & 13.68 & 44.18 & 39.90 & 21.89 & 18.96 & 86 & 74 & 77 & 34.24 & 4.73\\
+ Targeted Evidence Completion & 12.40 & 8.42 & 23.80 & 10.25 & 11.38 & 67.35 & 35 & 15.51 & 13.68 & 44.18 & 41.65 & 31.92 & 22.50 & 86 & 76 & 77 & 36.06 & 6.03\\
Query Crop                     & 2.10 & 4.49 & 16.33 & 5.55 & 4.22 & 58.16 & 37 & 13.77 & 6.58 & 23.40 & 36.68 & 12.80 & 14.56 & 75 & 76 & 74 & 28.79 & -\\
Sliding Window                 & 7.87 & 0.07 & 15.55 & 6.95 & 4.05 & 67.35 & 27 & 13.71 & 4.34 & 33.31 & 9.70 & 13.08 & 14.56 & 75 & 76 & 74 & 27.66 & 13.88 \\
\midrule
\multicolumn{19}{@{}l}{\textit{InternVL3.5-8B}} \\
\midrule
Baseline                       & 2.29 & 0.90 & 6.65 & 2.50 & 2.76 & 59.18 & 26 & 9.46 & 2.32 & 43.30 & 30.37 & 28.52 & 31.54 & 39 & 39 & 45 & 23.05 & 1.00\\
+ Evidence-centered Acquisition & 2.26 & 18.53 & 5.99 & 6.57 & 2.63 & 68.37 & 44 & 9.60 & 9.76 & 42.37 & 49.21 & 17.55 & 28.57 & 40 & 43 & 53 & 27.59 & 4.74 \\
+ Targeted Evidence Completion & 5.91 & 21.20 & 5.99 & 6.57 & 10.77 & 68.37 & 44 & 9.60 & 9.76 & 42.37 & 54.44 & 24.41 & 32.83 & 40 & 43 & 53 & 29.51 & 5.88 \\
Query Crop                     & 2.29 & 20.19 & 6.65 & 2.50 & 2.76 & 61.22 & 44 & 9.46 & 1.83 & 43.30 & 52.29 & 28.52 & 31.54 & 39 & 39 & 45 & 26.85 & -\\
Sliding Window                 & 2.29 & 2.33 & 4.99 & 5.49 & 2.58 & 66.33 & 36 & 15.98 & 2.68 & 38.30 & 42.67 & 17.02 & 31.54 & 39 & 39 & 45 & 24.45 & 13.88\\
\midrule
\multicolumn{19}{@{}l}{\textit{Zoom-based methods}} \\
\midrule
DeepEyes (Qwen2.5-VL-7B)       & 1.24 & 0.00 & 0.04 & 0.49 & 0.65 & 60.20 & 25 & 1.46 & 1.40 & 42.57 & 31.69 & 35.89 & 26.40 & 63 & 46 & 51 & 24.18 & 1.12\\
GeoEyes (Qwen2.5-VL-7B)        & 0.00 & 0.00 & 0.80 & 0.43 & 0.00 & 51.02 & 30 & 1.39 & 1.10 & 30.17 & 26.11 & 27.39 & 10.05 & 53 & 47 & 48 & 20.41 & 1.94\\
HiDe (Qwen2.5-VL-7B)           & 0.00 & 0.00 & 1.05 & 1.16 & 0.00 & 58.16 & 23 & 1.48 & 1.15 & 45.28 & 23.86 & 32.66 & 23.64 & 89 & 47 & 53 & 25.03 & 4.00\\
ZoomSearch (Qwen3-VL-8B)       & 0.00 & 0.00 & 0.15 & 0.17 & 0.00 & 48.98 & 26 & 0.63 & 0.01 & 32.67 & 4.11 & 18.00 & 12.50 & 59 & 46 & 55 & 18.95 & 10.50\\
ZoomEye (Qwen2.5-VL-3B)        & 0.00 & 0.00 & 0.47 & 1.43 & 0.00 & 58.16 & 27 & 0.00 & 0.07 & 21.91 & 28.36 & 42.02 & 30.35 & 24 & 43 & 37 & 19.61 & 129.06\\
\bottomrule
\end{tabular}
}
\vspace{-2mm}
\end{table*}

\subsection{Qualitative Examples of UHR-Micro}
\label{app:qualitative_examples}

\begin{figure}[h!]
\centering
\includegraphics[width=0.78\linewidth]{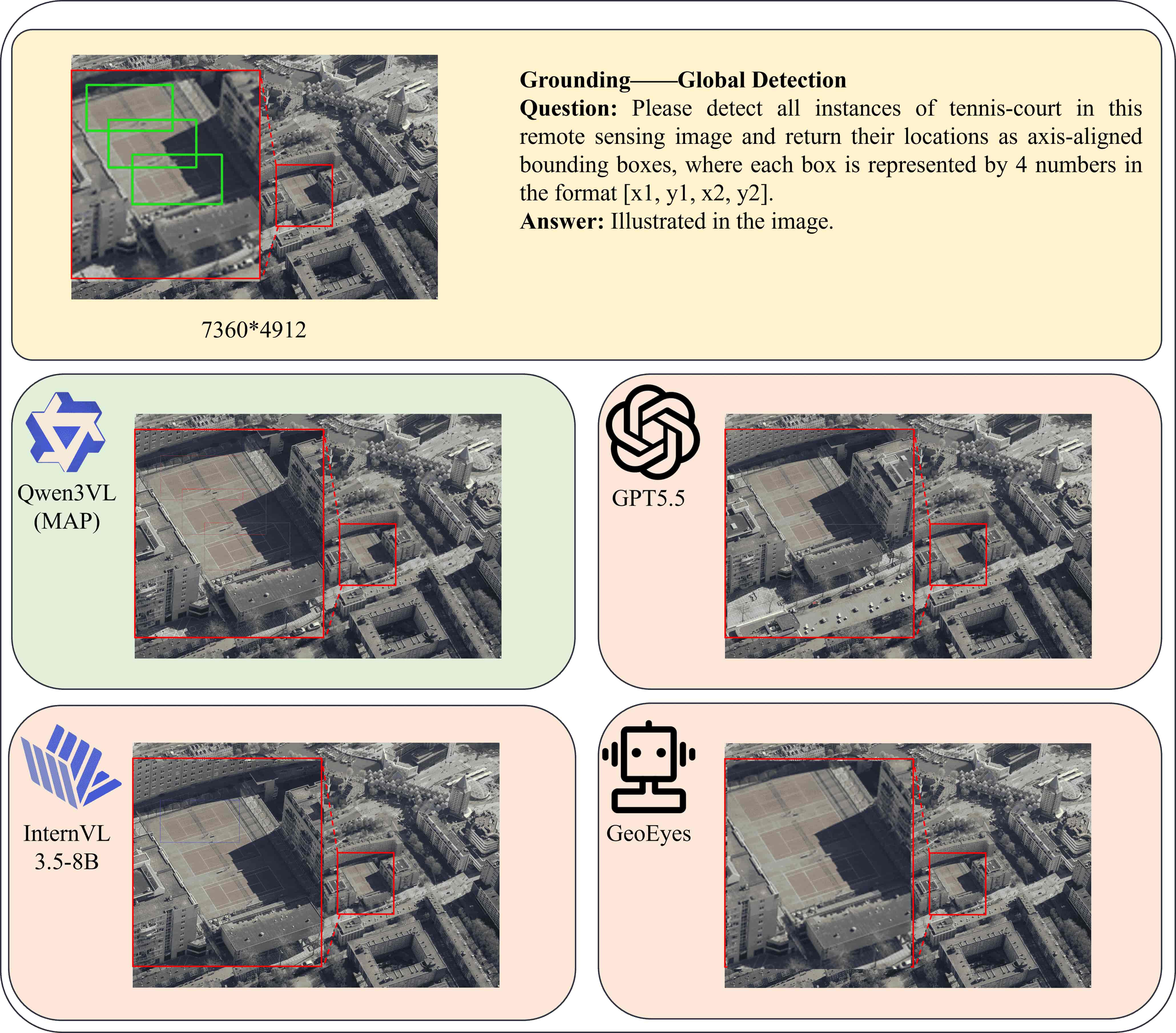}
\caption{
Example of \textit{Global Detection (GD)}. The figure shows the UHR image, instruction, ground-truth, and model predictions.
}
\label{fig:exa1}
\end{figure}

\begin{figure}[h!]
\centering
\includegraphics[width=0.78\linewidth]{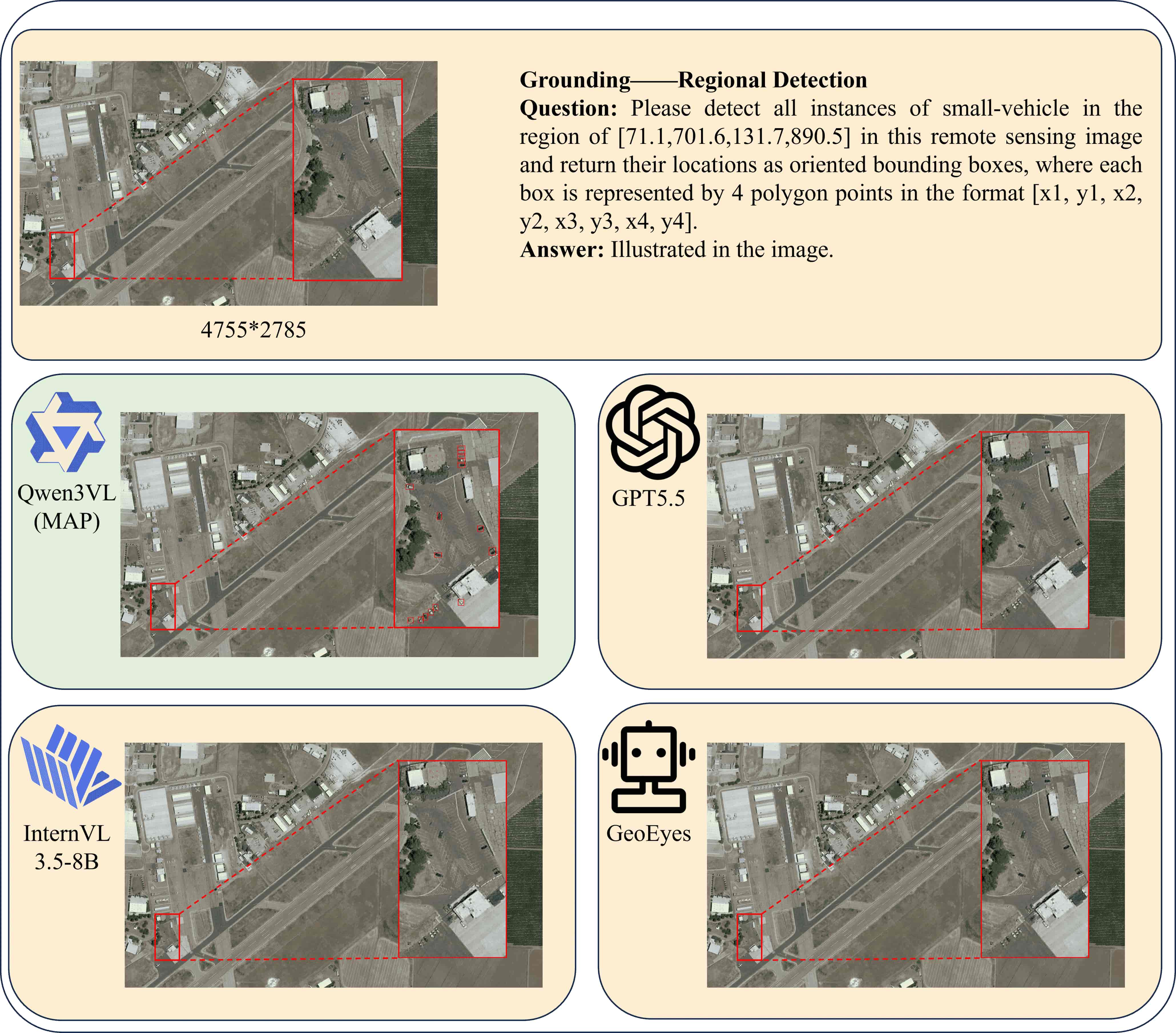}
\caption{
Example of \textit{Regional Detection (RD)}. The figure shows the UHR image, instruction, ground-truth, and model predictions.
}
\label{fig:exa2}
\end{figure}

\begin{figure}[h!]
\centering
\includegraphics[width=0.78\linewidth]{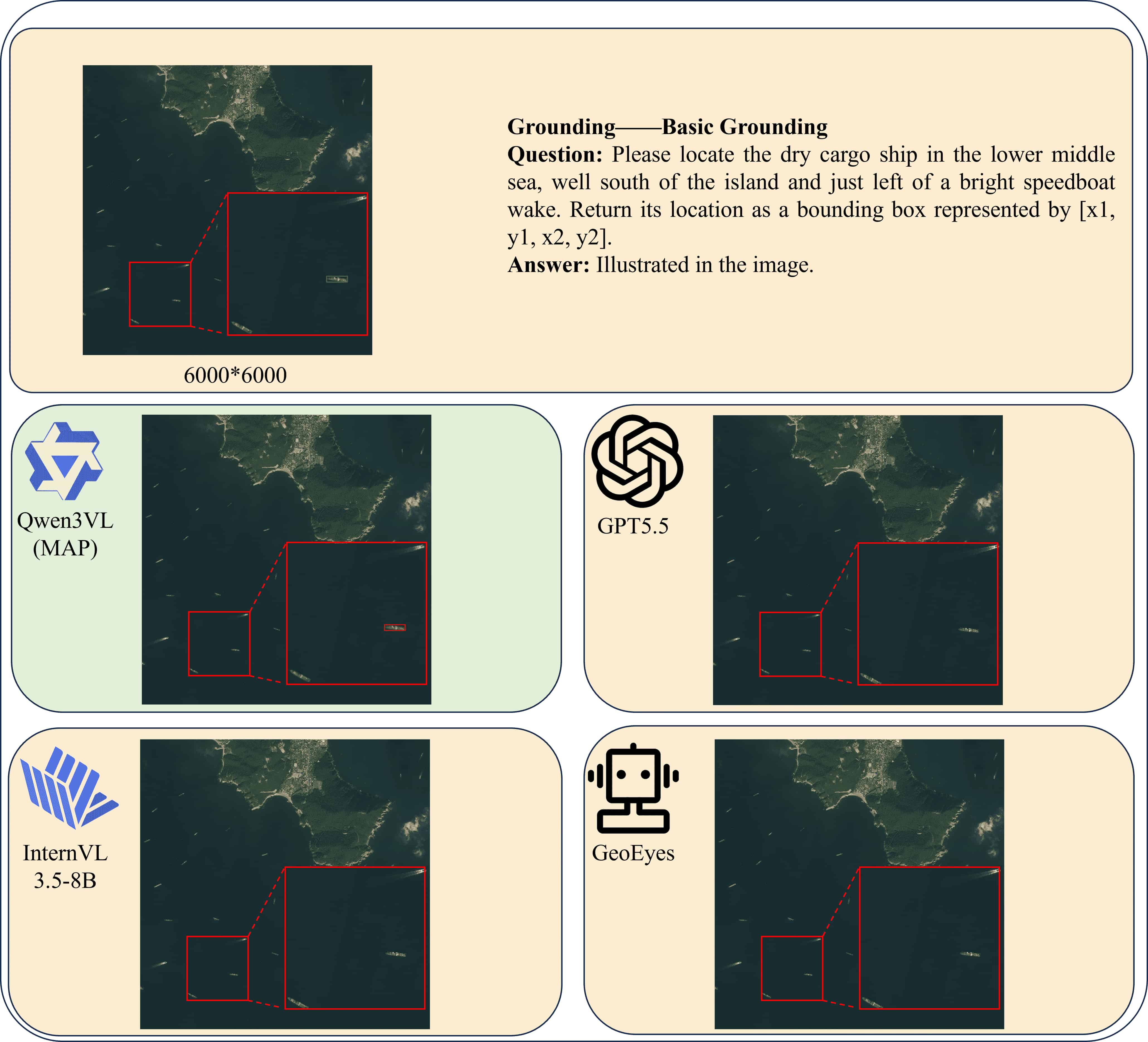}
\caption{
Example of \textit{Basic Grounding (BG)}. The figure shows the UHR image, instruction, ground-truth, and model predictions.
}
\label{fig:exa3}
\end{figure}

\begin{figure}[h!]
\centering
\includegraphics[width=0.78\linewidth]{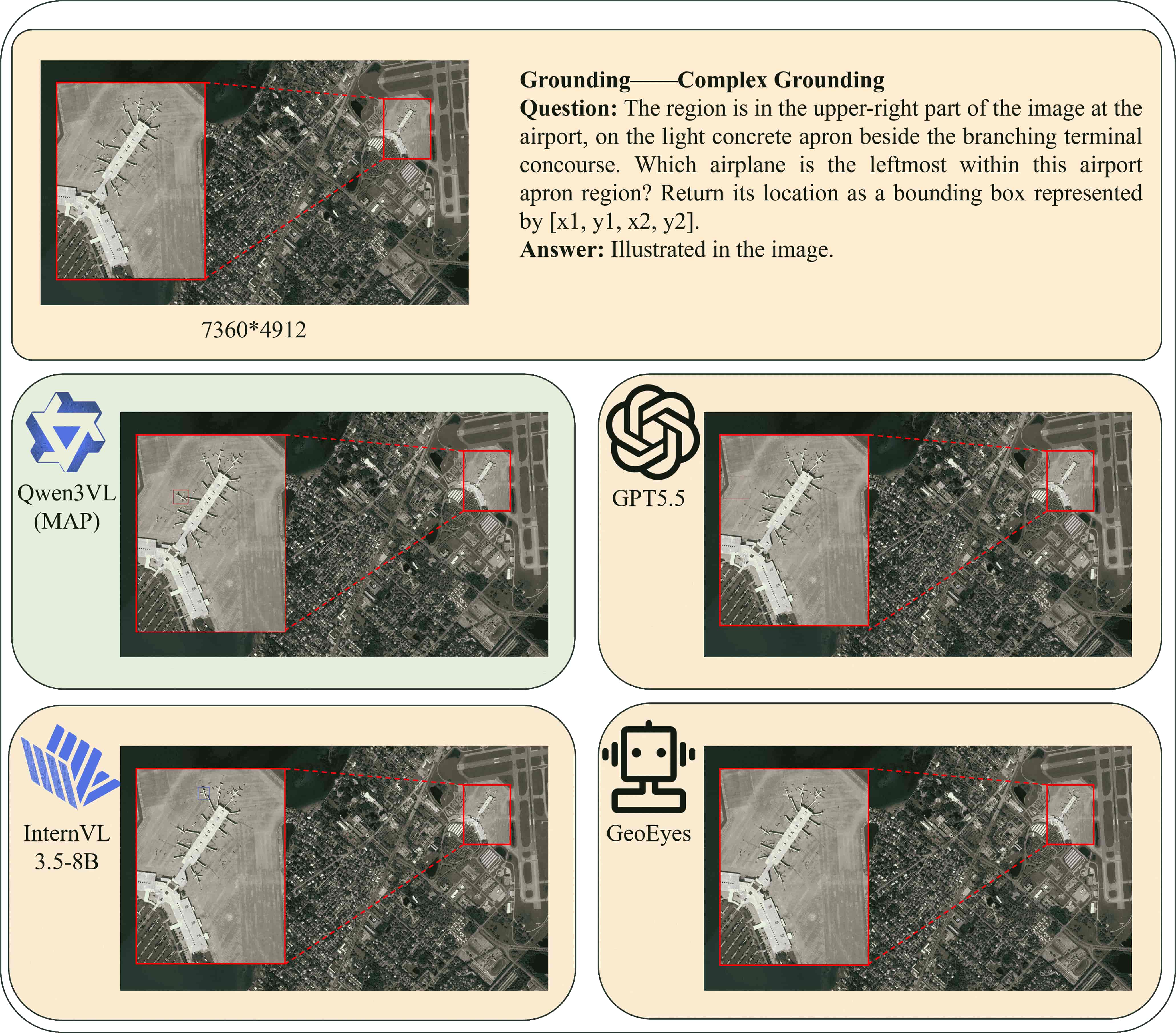}
\caption{
Example of \textit{Complex Grounding (CG)}. The figure shows the UHR image, instruction, ground-truth, and model predictions.
}
\label{fig:exa4}
\end{figure}

\begin{figure}[h!]
\centering
\includegraphics[width=0.78\linewidth]{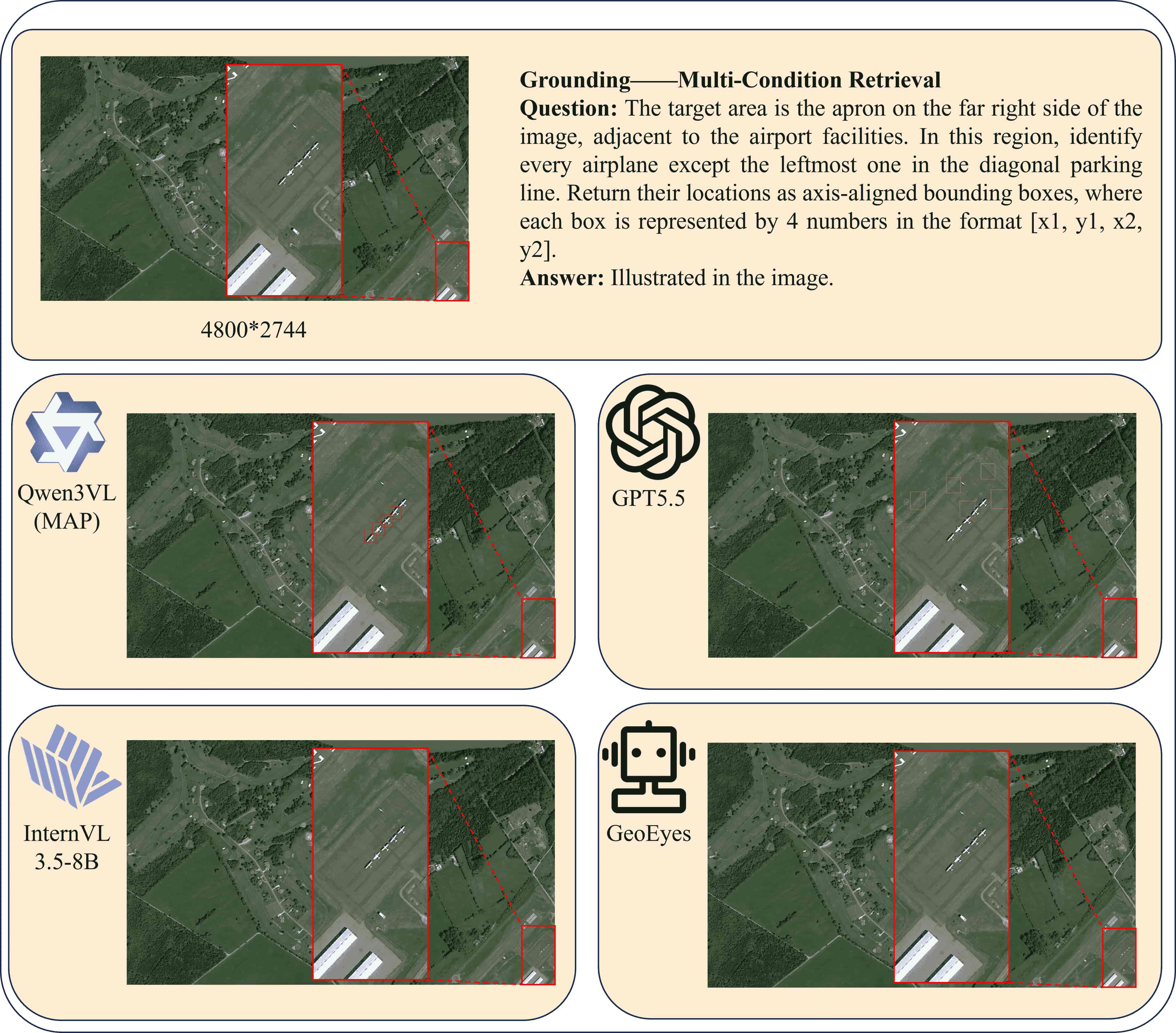}
\caption{
Example of \textit{Multi-Condition Retrieval (MCR)}. The figure shows the UHR image, instruction, ground-truth, and model predictions.
}
\label{fig:exa5}
\end{figure}

\begin{figure}[h!]
\centering
\includegraphics[width=0.78\linewidth]{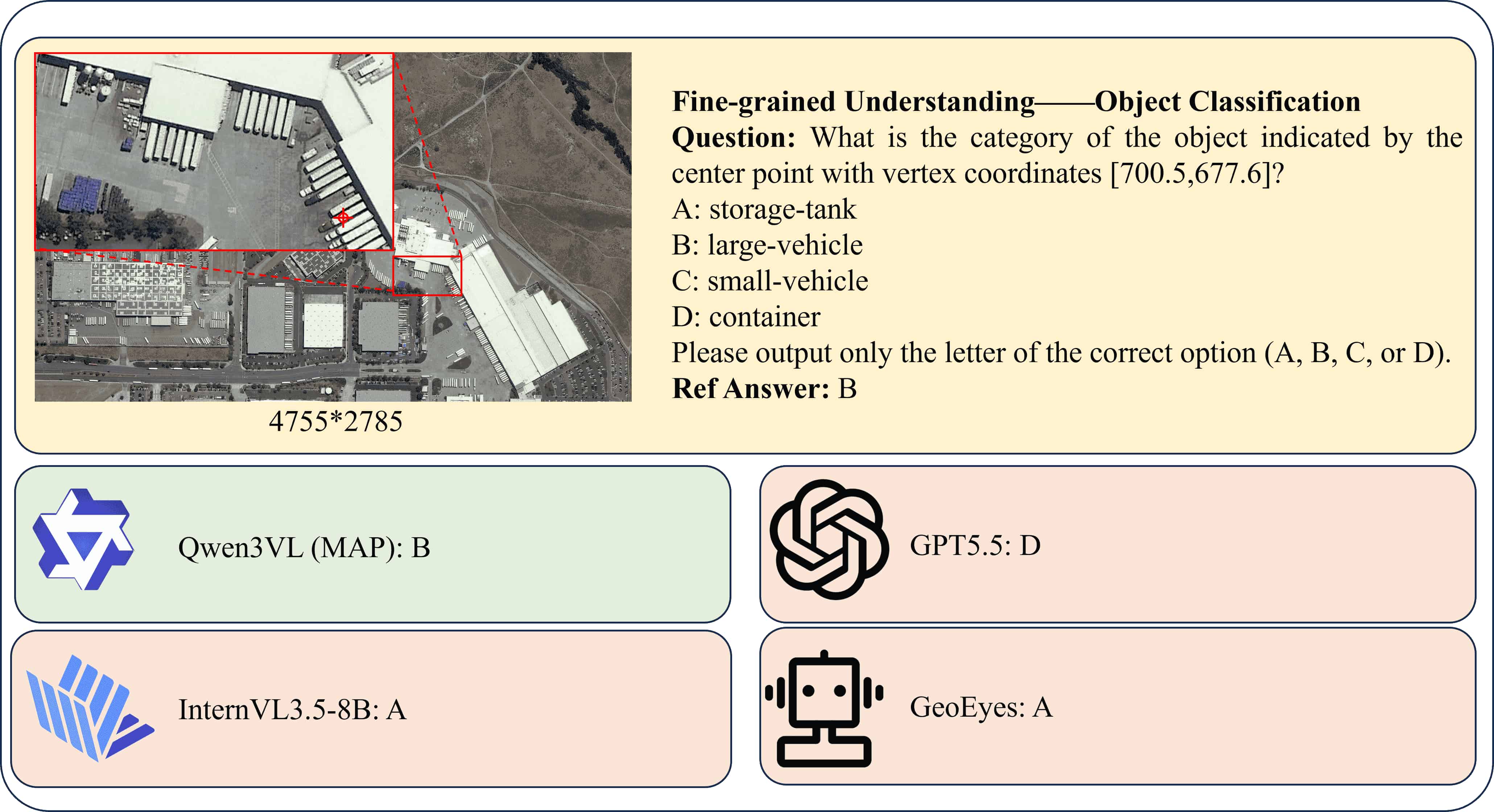}
\caption{
Example of \textit{Object Classification (OC)}. The figure shows the UHR image, instruction, ground-truth, and model predictions.
}
\label{fig:exa6}
\end{figure}

\begin{figure}[h!]
\centering
\includegraphics[width=0.78\linewidth]{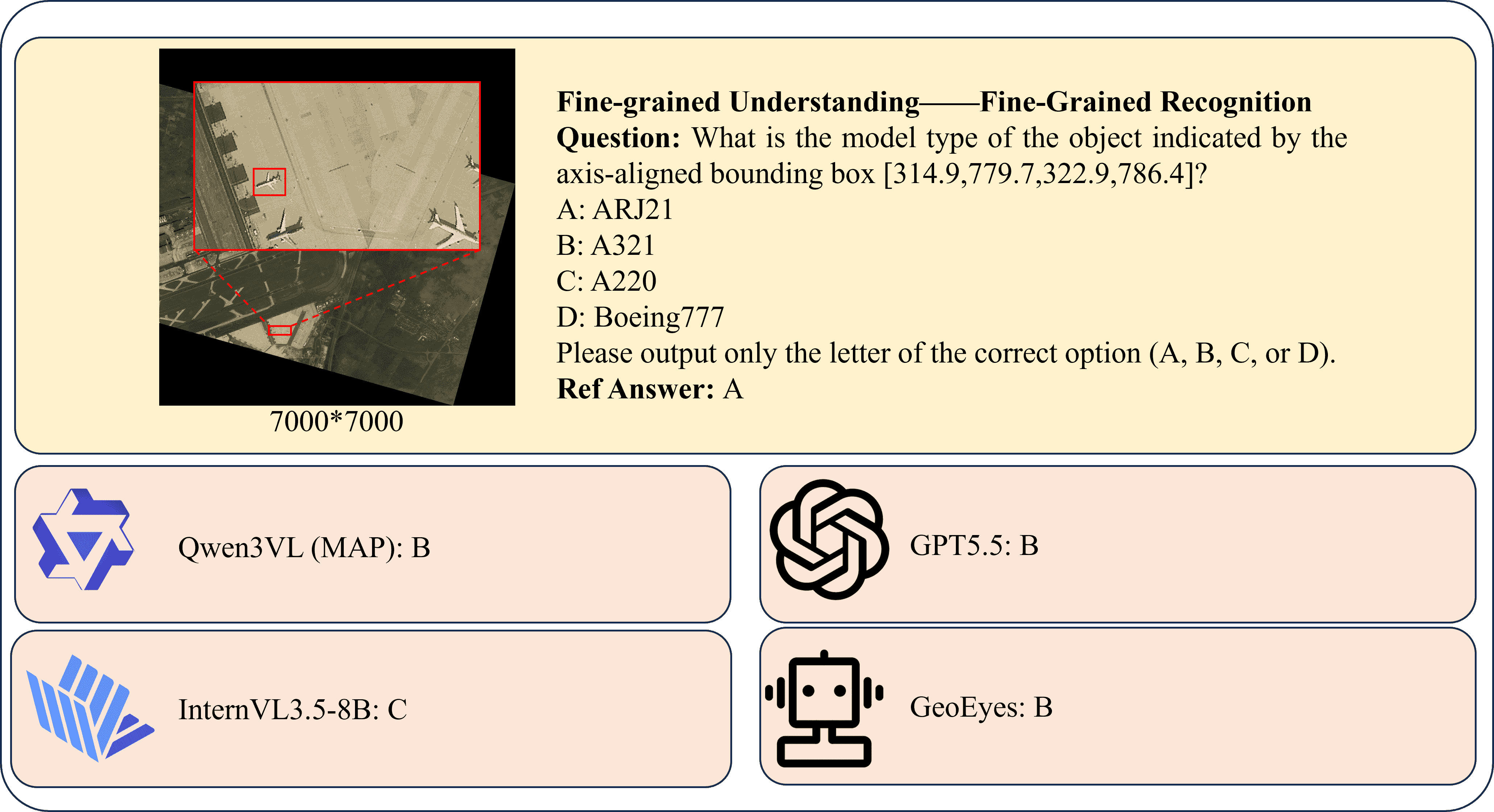}
\caption{
Example of \textit{Fine-Grained Recognition (FGR)}. The figure shows the UHR image, instruction, ground-truth, and model predictions.
}
\label{fig:exa7}
\end{figure}

\begin{figure}[h!]
\centering
\includegraphics[width=0.78\linewidth]{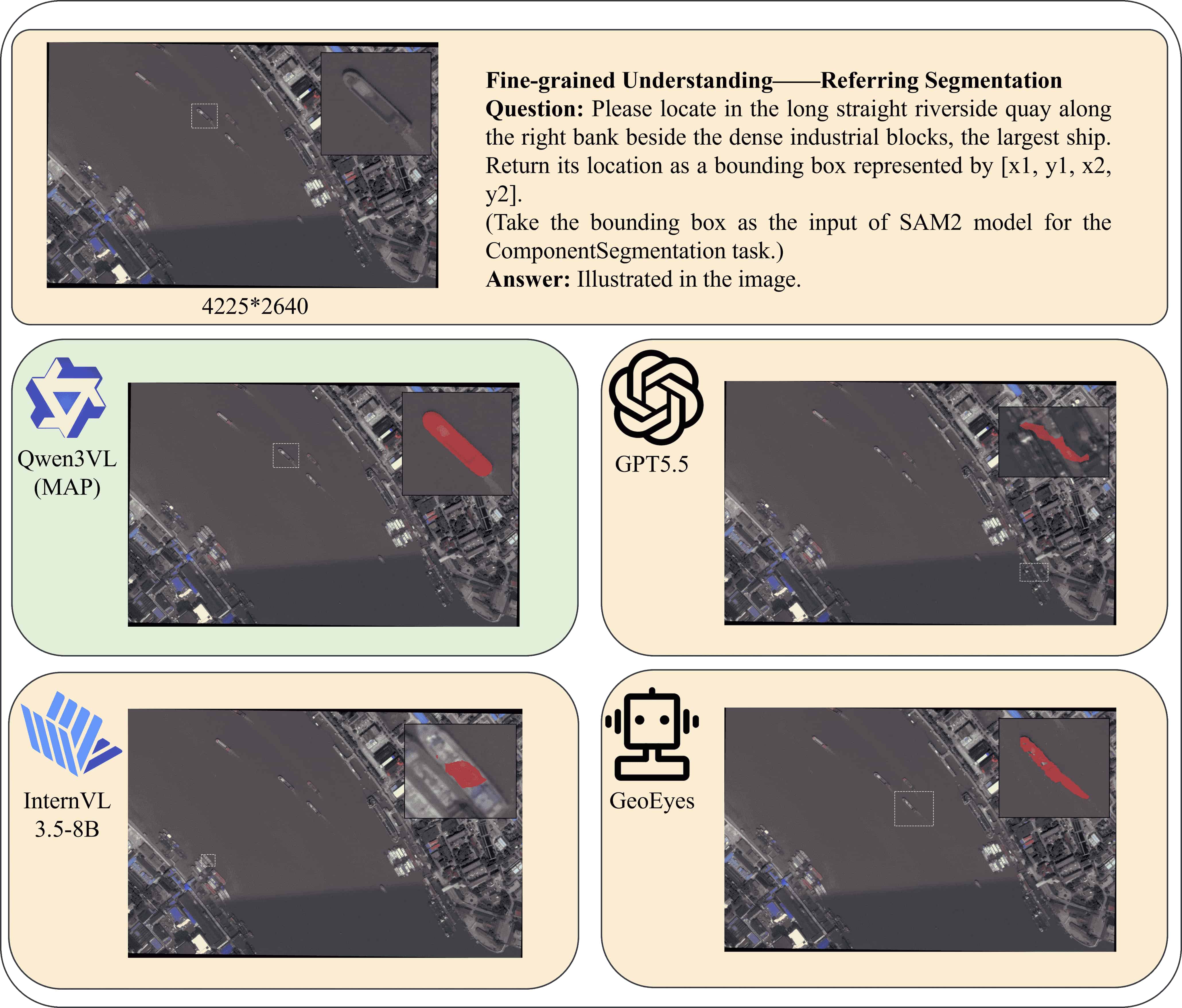}
\caption{
Example of \textit{Referring Segmentation (RS)}. The figure shows the UHR image, instruction, ground-truth, and model predictions.
}
\label{fig:exa8}
\end{figure}

\begin{figure}[h!]
\centering
\includegraphics[width=0.78\linewidth]{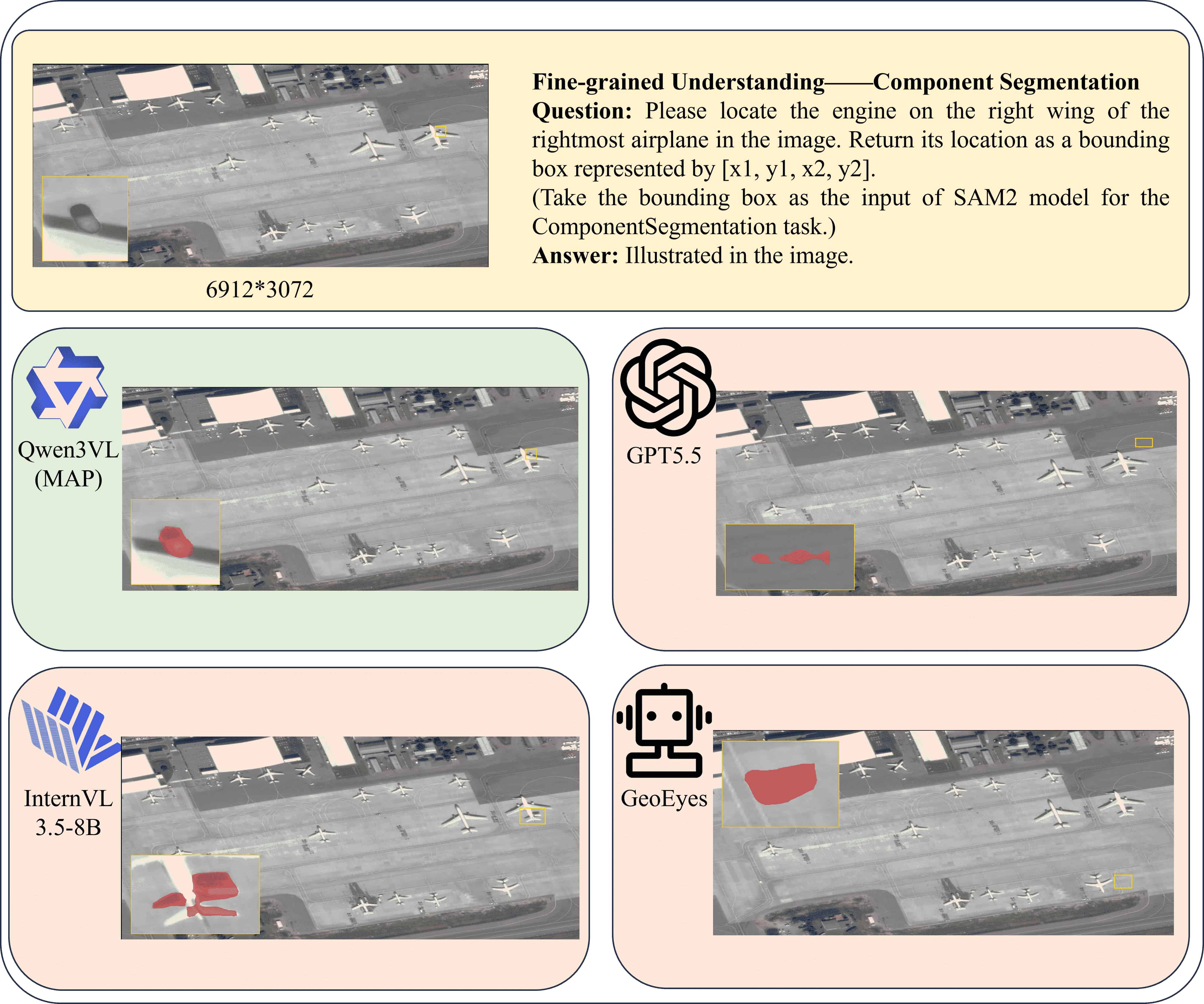}
\caption{
Example of \textit{Component Segmentation (CS)}. The figure shows the UHR image, instruction, ground-truth, and model predictions.
}
\label{fig:exa9}
\end{figure}

\begin{figure}[h!]
\centering
\includegraphics[width=0.78\linewidth]{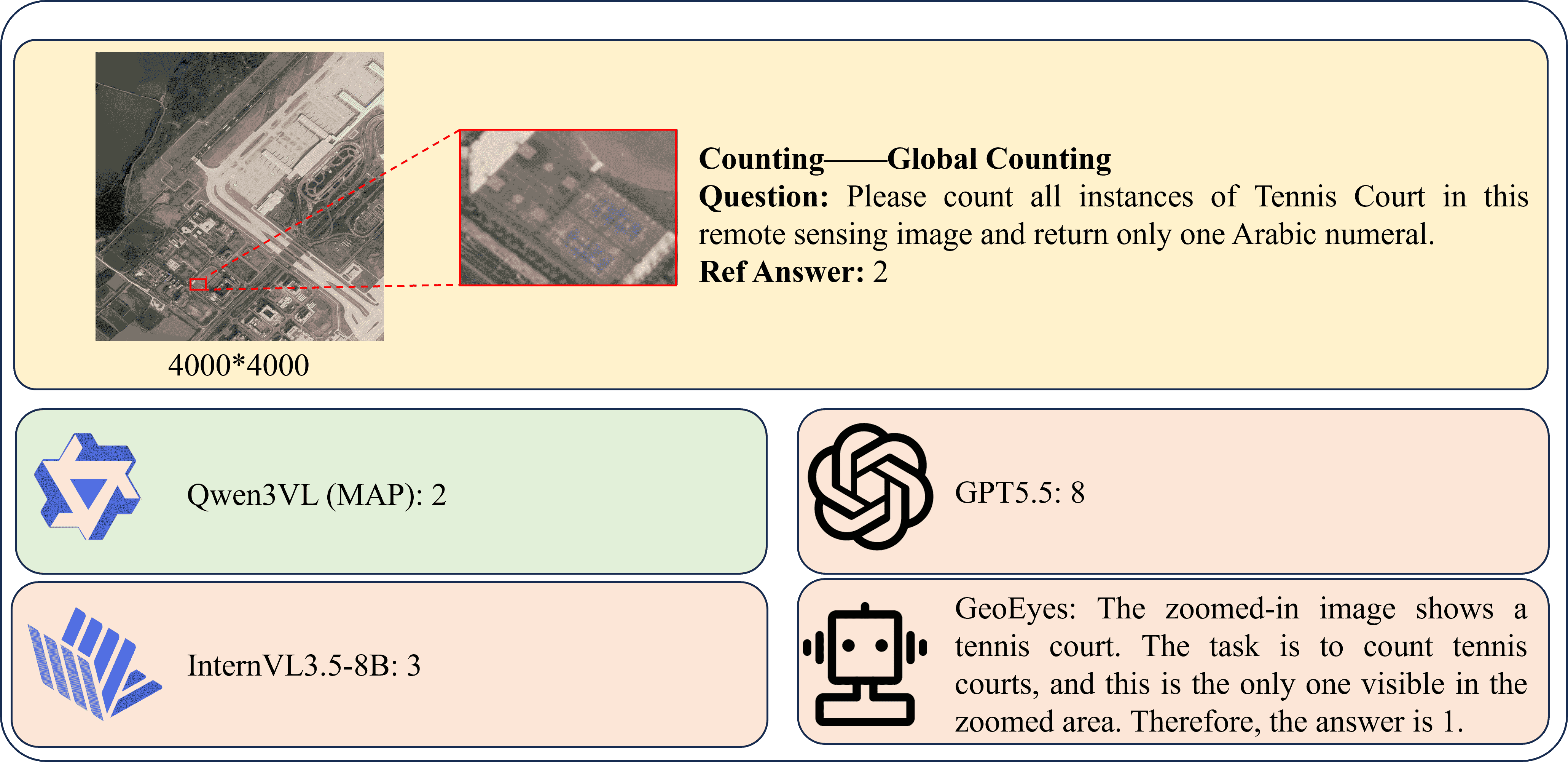}
\caption{
Example of \textit{Global Counting (GC)}. The figure shows the UHR image, instruction, ground-truth, and model predictions.
}
\label{fig:exa10}
\end{figure}

\begin{figure}[h!]
\centering
\includegraphics[width=0.78\linewidth]{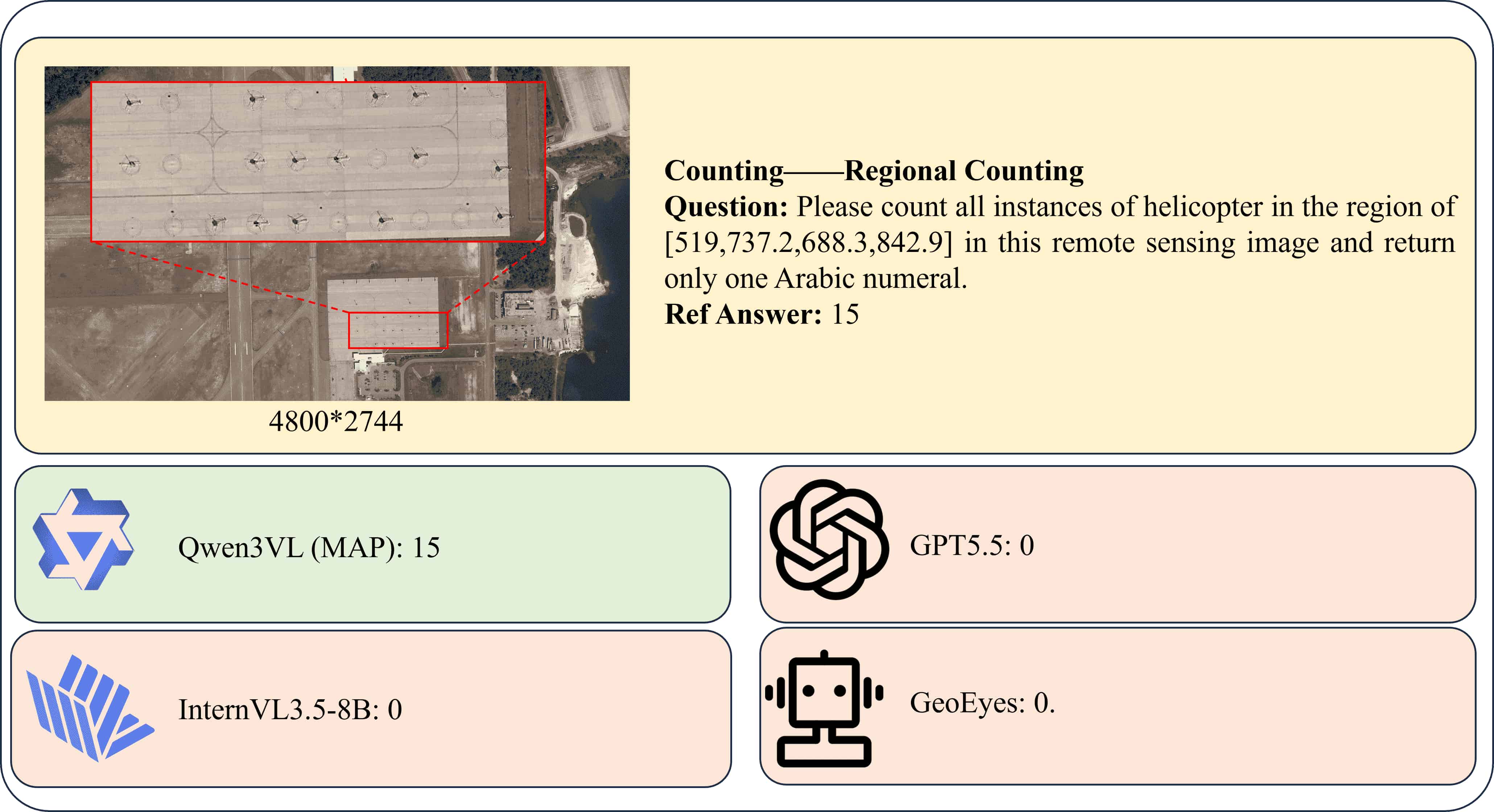}
\caption{
Example of \textit{Regional Counting (RC)}. The figure shows the UHR image, instruction, ground-truth, and model predictions.
}
\label{fig:exa11}
\end{figure}

\begin{figure}[h!]
\centering
\includegraphics[width=0.78\linewidth]{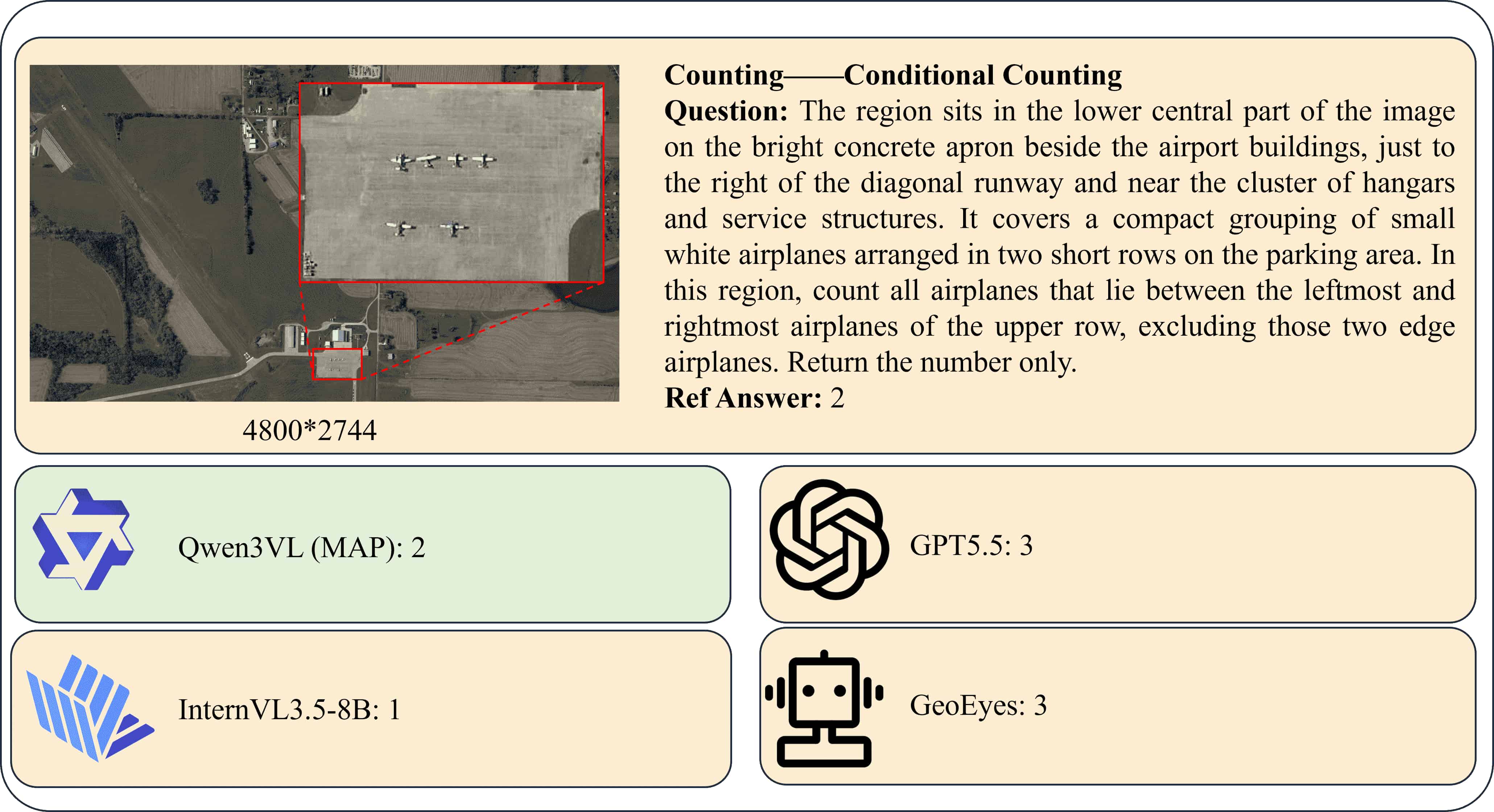}
\caption{
Example of \textit{Conditional Counting (CC)}. The figure shows the UHR image, instruction, ground-truth, and model predictions.
}
\label{fig:exa12}
\end{figure}

\begin{figure}[h!]
\centering
\includegraphics[width=0.78\linewidth]{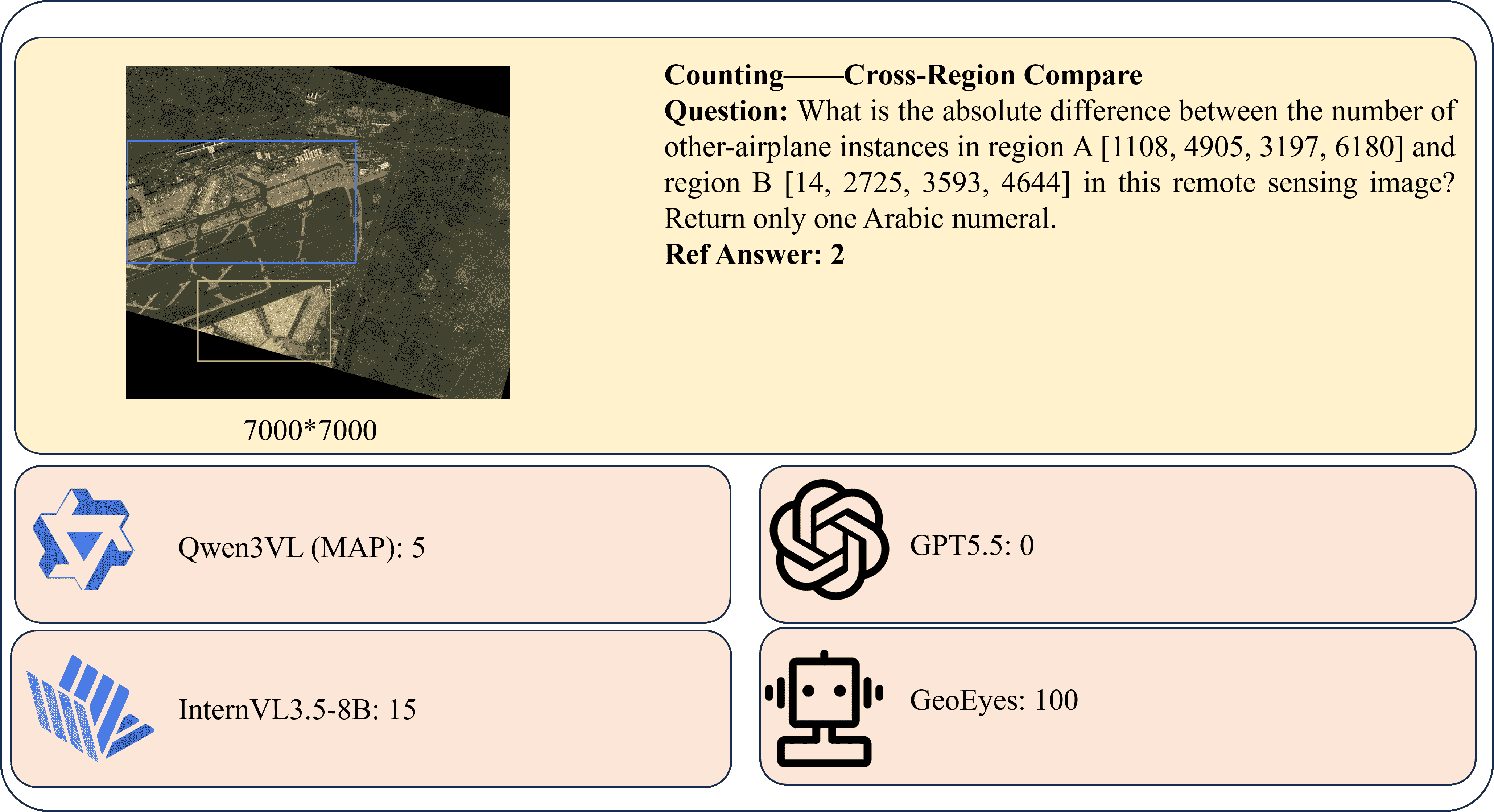}
\caption{
Example of \textit{Cross-Region Compare (CRC)}. The figure shows the UHR image, instruction, ground-truth, and model predictions.
}
\label{fig:exa13}
\end{figure}

\begin{figure}[h!]
\centering
\includegraphics[width=0.78\linewidth]{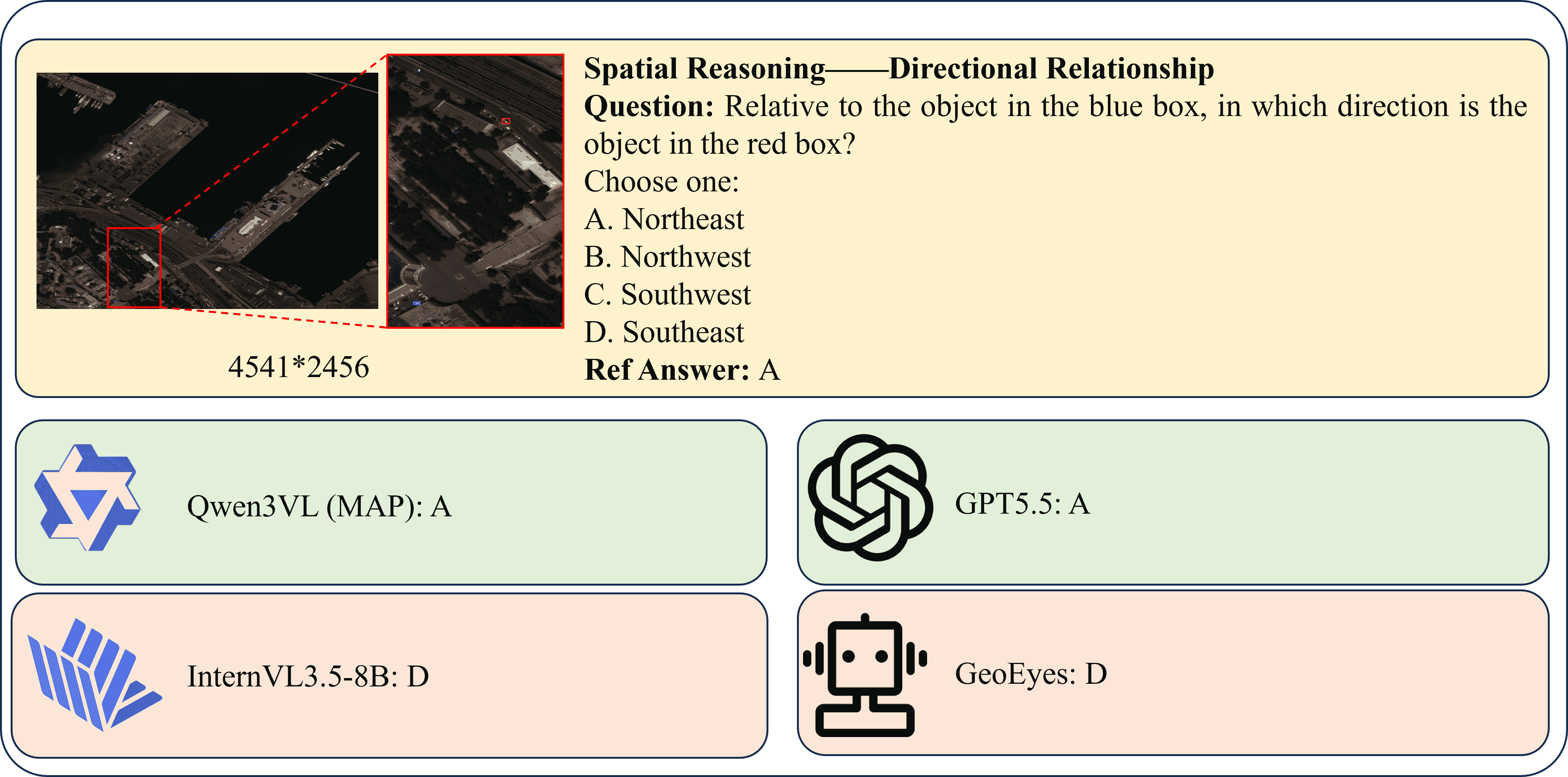}
\caption{
Example of \textit{Directional Relationship (DrR)}. The figure shows the UHR image, instruction, ground-truth, and model predictions.
}
\label{fig:exa14}
\end{figure}

\begin{figure}[h!]
\centering
\includegraphics[width=0.78\linewidth]{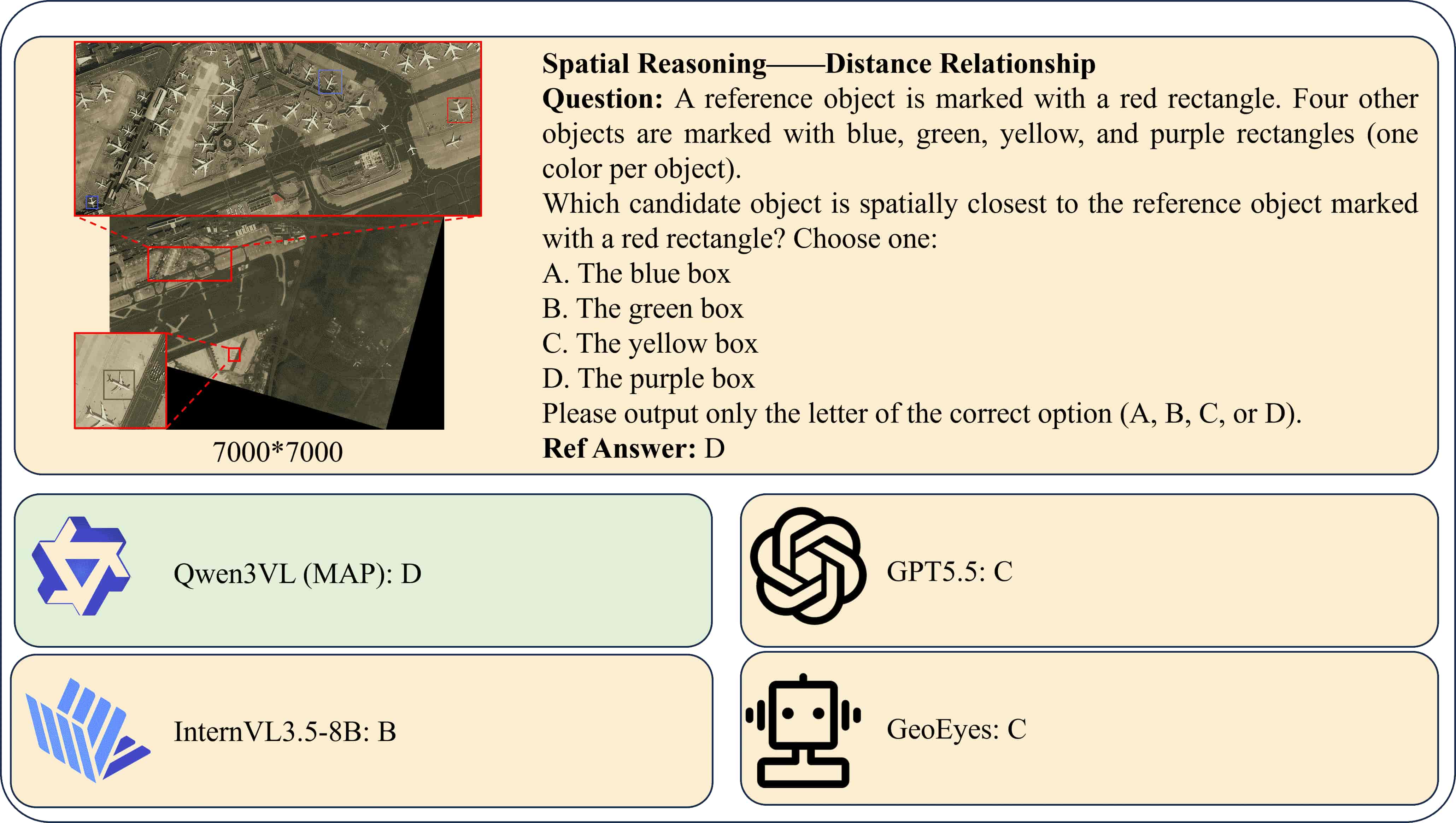}
\caption{
Example of \textit{Distance Relationship (DsR)}. The figure shows the UHR image, instruction, ground-truth, and model predictions.
}
\label{fig:exa15}
\end{figure}

\begin{figure}[h!]
\centering
\includegraphics[width=0.78\linewidth]{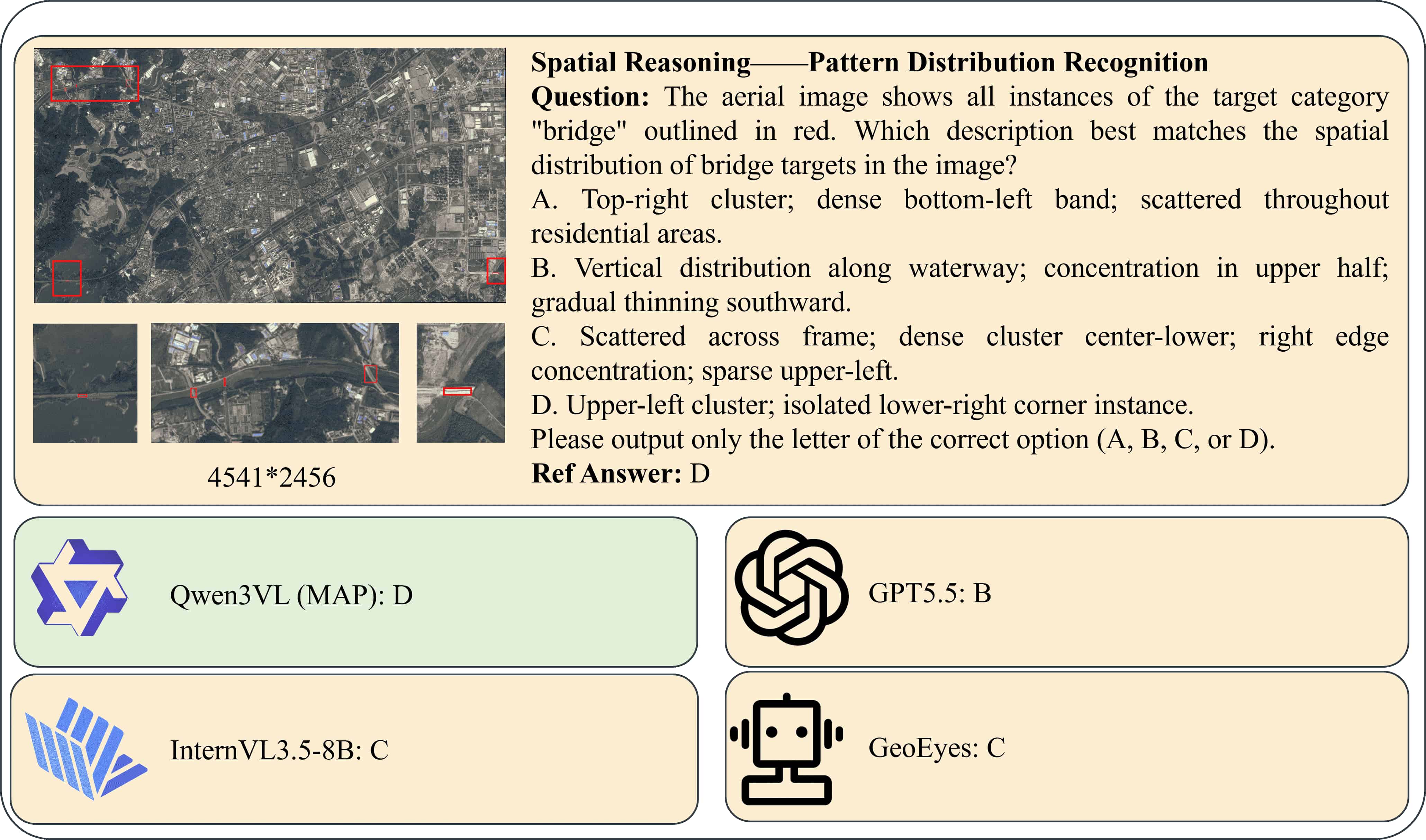}
\caption{
Example of \textit{Pattern Distribution Recognition (PDR)}. The figure shows the UHR image, instruction, ground-truth, and model predictions.
}
\label{fig:exa16}
\end{figure}

This section presents qualitative examples from UHR-Micro to illustrate the diversity of task formulations, visual conditions, answer formats, and model behaviors. We provide one representative example for each of the 16 tasks in the benchmark, as shown in Fig.~\ref{fig:exa1}--\ref{fig:exa16}. Each example contains the UHR image or the relevant visual crop, the natural-language instruction, the ground-truth answer, and predictions from representative models. These examples complement the quantitative results by showing how micro-target perception failures arise in realistic UHR scenes.

\subsection{Datasheets}
\label{app:datasheets}

\subsubsection{Motivation}

\begin{enumerate}
    \item \textit{``For what purpose was the dataset created?''}
    
    \textcolor{BurntOrange}{\textbf{A:}} 
UHR-Micro was created to address the lack of systematic benchmarks for evaluating micro-target perception in ultra-high-resolution Earth observation imagery. Existing multimodal remote sensing benchmarks primarily emphasize scene-level semantics, salient objects, or cropped image patches, leaving it unclear whether vision--language models can locate and reason over extremely small but task-critical visual evidence within native UHR scenes. UHR-Micro fills this gap by providing a diagnostic evaluation suite with deterministic instructions across grounding, fine-grained understanding, counting, and spatial reasoning tasks. The dataset is designed to expose the scale disparity between tiny targets and massive image canvases, operationalize the resulting resolution illusion, and support controlled analysis of failure modes such as missed localization, hallucination, and coordinate shift. Overall, UHR-Micro aims to facilitate the development and rigorous benchmarking of perception-aware VLMs capable of reliable global-to-local reasoning in real-world UHR imagery.
    
    \item \textit{``Who created the dataset (\textit{e.g.}, which team, research group) and on behalf of which entity?''}
    
    \textcolor{BurntOrange}{\textbf{A:}} The dataset was created by the following authors:
    \begin{itemize}
      \item Anonymous authors
    \end{itemize}
    
    \item \textit{``Who funded the creation of the dataset?''}
    
    \textcolor{BurntOrange}{\textbf{A:}}
    The dataset creation was funded by the affiliations of the authors involved in this work.
\end{enumerate}

\subsubsection{Composition}
Most of the questions in this section are intended to provide dataset consumers with the information they need to make informed decisions about using the dataset for their chosen tasks. Some of the questions are designed to elicit information about compliance with the EU’s General Data Protection Regulation (GDPR) or comparable regulations in other jurisdictions. Questions that apply only to datasets that relate to people are grouped together at the end of the section. We recommend taking a broad interpretation of whether a dataset relates to people. For example, any dataset containing text that was written by people relates to people.
\begin{enumerate}
    \item \textit{``What do the instances that comprise our datasets represent (\textit{e.g.}, documents, photos, people, countries)?''}
    
    \textcolor{BurntOrange}{\textbf{A:}} The dataset primarily consists of remote sensing images captured by satellites and drones, along with task-specific textual annotations. The source datasets used in UHR-Micro are publicly released remote sensing datasets; UHR-Micro preserves source attribution and follows the corresponding source licenses and terms.
    
    \item \textit{``How many instances are there in total (of each type, if appropriate)?''}
    
    \textcolor{BurntOrange}{\textbf{A:}} UHR-Micro includes 11,072 instructions over 1,212 UHR images. Details can be found in the main text.

    \item \textit{``Does the dataset contain all possible instances or is it a sample (not necessarily random) of instances from a larger set?''}
    
    \textcolor{BurntOrange}{\textbf{A:}} The images in UHR-Micro are sourced from existing detection and segmentation  datasets. All textual annotations were independently created by us.
    
    \item \textit{``Is there a label or target associated with each instance?''}
    
    \textcolor{BurntOrange}{\textbf{A:}} Yes. Each instance contains an image, an instruction, and a deterministic target answer.
    
    \item \textit{``Is any information missing from individual instances?''}
    
    \textcolor{BurntOrange}{\textbf{A:}} No, each individual instance is complete.
    
    \item \textit{``Are relationships between individual instances made explicit (\textit{e.g.}, users’ movie ratings, social network links)?''}
    
    \textcolor{BurntOrange}{\textbf{A:}} Yes. Each instruction is associated with its source image and task-specific target annotations; no social or user-level relationships are included.
    
    \item \textit{``Are there recommended data splits (\textit{e.g.}, training, development/validation, testing)?''}
    
    \textcolor{BurntOrange}{\textbf{A:}} 
    Yes, UHR-Micro provides development, validation, and test query splits.
    
    \item \textit{``Is the dataset self-contained, or does it link to or otherwise rely on external resources (\textit{e.g.}, websites, tweets, other datasets)?''}
    
    \textcolor{BurntOrange}{\textbf{A:}} UHR-Micro will be released with processed images, annotations, source metadata, and usage documentation on platforms such as GitHub and Hugging Face, subject to the licenses and terms of the underlying source datasets.
    
    \item \textit{``Does the dataset contain data that might be considered confidential (\textit{e.g.}, data that is protected by legal privilege or by doctor–patient confidentiality, data that includes the content of individuals’ non-public communications)?''}
    
    \textcolor{BurntOrange}{\textbf{A:}} To the best of our knowledge, UHR-Micro does not contain confidential records, private communications, or medical, financial, or legal records. The images are derived from publicly released remote sensing datasets, and annotations are constructed for benchmark evaluation.
    
    \item \textit{``Does the dataset contain data that, if viewed directly, might be offensive, insulting, threatening, or might otherwise cause anxiety?''}
    
    \textcolor{BurntOrange}{\textbf{A:}} UHR-Micro does not intentionally include offensive, insulting, or graphic content. However, as an Earth observation benchmark, it may contain publicly observable infrastructure, vehicles, vessels, or facilities, and should therefore be used with appropriate awareness of dual-use risks.
\end{enumerate}

\subsubsection{Collection Process}
In addition to the goals outlined in the previous section, the questions in this section are designed to elicit information that may help researchers and practitioners create alternative datasets with similar characteristics. Again, questions that apply only to datasets that relate to people are grouped together at the end of the section.
\begin{enumerate}
    \item \textit{``How was the data associated with each instance acquired?''}
    
    \textcolor{BurntOrange}{\textbf{A:}} 
The images in UHR-Micro are sourced from existing public remote sensing detection and segmentation datasets, and we enrich them with task-specific textual instructions and target answers. Details are provided in the main text.
    
    \item \textit{``What mechanisms or procedures were used to collect the data (\textit{e.g.}, hardware apparatuses or sensors, manual human curation, software programs, software APIs)?''}
    
    \textcolor{BurntOrange}{\textbf{A:}} 
UHR-Micro was constructed by curating images and annotations from established public remote sensing datasets. No additional hardware apparatuses, sensors, or image acquisition campaigns were introduced. We first filtered source images to retain 4K-scale ultra-high-resolution scenes with sufficient dense object annotations. Based on these native UHR images and their ground-truth annotations, we built a three-tier hybrid data engine. First, foundational tasks driven by coordinate, category, counting, and spatial-relation mappings were generated programmatically using geometric rules and existing dense annotations. Second, tasks requiring descriptive semantic alignment, such as basic grounding and pattern distribution recognition, were generated with the assistance of a frontier vision--language model through API-based prompting and then cross-validated by human annotators. Third, tasks involving complex multi-hop spatial reasoning, including complex grounding, cross-region comparative counting, and multi-condition retrieval, were constructed under a human-led protocol by domain experts, with model assistance used only as auxiliary support. All candidate samples were subjected to format validation, target-existence checks, ambiguity filtering, and expert review where necessary to ensure deterministic and unambiguous evaluation targets.
    
    \item \textit{``If the dataset is a sample from a larger set, what was the sampling strategy (\textit{e.g.}, deterministic, probabilistic with specific sampling probabilities)?''} 
    
    \textcolor{BurntOrange}{\textbf{A:}} Please refer to the details listed in the main text Section 3.
\end{enumerate}

\subsubsection{Preprocessing, Cleaning, and Labeling}
The questions in this section are intended to provide dataset
consumers with the information they need to determine whether the “raw” data has been processed in ways that are compatible with their chosen tasks. For example, text that has been converted into a ``bag-of-words" is not suitable for tasks involving word order.
\begin{enumerate}
    \item \textit{``Was any preprocessing/cleaning/labeling of the data done (\textit{e.g.}, discretization or bucketing, tokenization, part-of-speech tagging, SIFT feature extraction, removal of instances, processing of missing values)?''}
    
    \textcolor{BurntOrange}{\textbf{A:}} 
Yes. UHR-Micro involved preprocessing, cleaning, and labeling steps to ensure that all samples support deterministic evaluation. During preprocessing, source annotations from different remote sensing datasets were converted into a unified format, and 4K-scale UHR images with valid dense object annotations were retained. Corrupted images, incomplete annotations, ambiguous targets, and samples whose answers could not be deterministically verified were removed. For task construction, existing object annotations were transformed into task-specific labels, including bounding boxes, segmentation masks, numerical answers, and discrete options. Coordinate-based tasks, counting tasks, and spatial-relation tasks were generated through rule-based geometric processing, while semantically richer tasks were labeled through VLM-assisted generation followed by human verification or expert-led annotation. All instructions and answers were further checked for format consistency, target existence, duplicate or near-duplicate cases, and ambiguity before being included in the final development, validation, and test splits.

    \item \textit{``Was the `raw' data saved in addition to the preprocessed/cleaned/labeled data (\textit{e.g.}, to support unanticipated future uses)?''} 
    
    \textcolor{BurntOrange}{\textbf{A:}} Yes. The underlying source images and annotations remain accessible through their original public dataset releases, subject to their respective licenses and terms.
    
    \item \textit{``Is the software that was used to preprocess/clean/label the data available?''} 
    
    \textcolor{BurntOrange}{\textbf{A:}} Yes. The preprocessing and evaluation scripts will be made publicly available with the benchmark release.
\end{enumerate}

\subsubsection{Uses}
The questions in this section are intended to encourage dataset creators to reflect on tasks for which the dataset should and should not be used. By explicitly highlighting these tasks, dataset creators can help dataset consumers make informed decisions, thereby avoiding potential risks or harms.
\begin{enumerate}
    \item \textit{``Has the dataset been used for any tasks already?''} 
    
    \textcolor{BurntOrange}{\textbf{A:}} 
    Yes. It is used in this paper to benchmark micro-target perception and global-to-local reasoning in VLMs.
    
    \item \textit{``Is there a repository that links to any or all papers or systems that use the dataset?''} 
    
    \textcolor{BurntOrange}{\textbf{A:}} Yes. We will provide such links on the project GitHub and Hugging Face repositories.
    
    \item \textit{``What (other) tasks could the dataset be used for?''}
    
    \textcolor{BurntOrange}{\textbf{A:}} 
Beyond benchmarking micro-target perception in VLMs, UHR-Micro can support research on high-resolution object detection, referring expression grounding, fine-grained remote sensing recognition, dense object counting, spatial relationship reasoning, active visual search, and agentic global-to-local perception.
    
    \item \textit{``Is there anything about the composition of the dataset or the way it was collected and preprocessed/cleaned/labeled that might impact future uses?''} 
    
    \textcolor{BurntOrange}{\textbf{A:}} No.
    
    \item \textit{``Are there tasks for which the dataset should not be used?''} 
    
    \textcolor{BurntOrange}{\textbf{A:}} UHR-Micro should not be used for privacy-invasive monitoring, operational surveillance, targeting, or safety-critical decision-making without appropriate validation, authorization, and human oversight.
\end{enumerate}

\subsubsection{Distribution}
Dataset creators should provide answers to these questions prior to distributing the dataset either internally within the entity on behalf of which the dataset was created or externally to third parties.
\begin{enumerate}
    \item \textit{``Will the dataset be distributed to third parties outside of the entity (\textit{e.g.}, company, institution, organization) on behalf of which the dataset was created?''} 
    
    \textcolor{BurntOrange}{\textbf{A:}} The benchmark will be made publicly accessible to the research community, subject to the licenses and terms of the underlying source datasets.
    
    \item \textit{``How will the dataset be distributed (\textit{e.g.}, tarball on website, API, GitHub)?''} 
    
    \textcolor{BurntOrange}{\textbf{A:}} We will provide UHR-Micro in the GitHub and the Hugging face repository.
    
    \item \textit{``When will the dataset be distributed?''} 
    
    \textcolor{BurntOrange}{\textbf{A:}} We will create a repository to release the data once the paper is officially published.
    
    \item \textit{``Will the dataset be distributed under a copyright or other intellectual property (IP) license, and/or under applicable terms of use (ToU)?''} 
    
    \textcolor{BurntOrange}{\textbf{A:}} Yes. The newly created UHR-Micro instructions, annotations, and metadata will be released under the Creative Commons Attribution-NonCommercial-ShareAlike 4.0 International License, while source imagery remains subject to the licenses and terms of the original datasets.
    
    \item \textit{``Have any third parties imposed IP-based or other restrictions on the data associated with the instances?''} 
    
    \textcolor{BurntOrange}{\textbf{A:}} Yes. The underlying source images are subject to the licenses and terms of their original datasets. UHR-Micro will preserve source attribution and document applicable usage constraints in the release.
    
    \item \textit{``Do any export controls or other regulatory restrictions apply to the dataset or to individual instances?''} 
    
    \textcolor{BurntOrange}{\textbf{A:}} We are not aware of export-control restrictions specific to UHR-Micro beyond the licenses and terms of the underlying source datasets. Users are responsible for ensuring compliance with applicable local regulations.    
\end{enumerate}

\subsubsection{Maintenance}
As with the questions in the previous section, dataset creators should provide answers to these questions prior to distributing the dataset. The questions in this section are intended to encourage dataset creators to plan for dataset maintenance and communicate this plan to dataset consumers.
\begin{enumerate}
    \item \textit{``Who will be supporting/hosting/maintaining the dataset?''} 
    
    \textcolor{BurntOrange}{\textbf{A:}} The authors of this work will support, host, and maintain the dataset.
    
    \item \textit{``How can the owner/curator/manager of the dataset be contacted (\textit{e.g.}, email address)?''} 
    
    \textcolor{BurntOrange}{\textbf{A:}} They can be contacted via the email addresses listed on the paper or webpage.
    
    \item \textit{``Is there an erratum?''} 
    
    \textcolor{BurntOrange}{\textbf{A:}} There is no erratum for the initial release; updates and known issues will be documented on the project repository.
    
    \item \textit{``Will the dataset be updated (\textit{e.g.}, to correct labeling errors, add new instances, delete instances)?''} 
    
    \textcolor{BurntOrange}{\textbf{A:}} Future updates (if any) will be posted on the dataset website.
    
    \item \textit{``Will older versions of the dataset continue to be supported/hosted/maintained?''} 
    
    \textcolor{BurntOrange}{\textbf{A:}} 

    Yes. Older versions will be archived when feasible, and updates will be documented on the project repository.
    
    \item \textit{``If others want to extend/augment/build on/contribute to the dataset, is there a mechanism for them to do so?''} 
    
    \textcolor{BurntOrange}{\textbf{A:}} Yes, we will provide detailed instructions for future extensions.
\end{enumerate}

\end{document}